\documentclass[runningheads]{llncs}
\usepackage{graphicx}
\usepackage{times}
\usepackage{xcolor}
\usepackage{soul}
\usepackage[utf8]{inputenc}
\usepackage[small]{caption}

\usepackage{algorithm}
\usepackage{algpseudocode}
\usepackage{subfig}
\usepackage[euler]{textgreek}
\usepackage{amssymb}
\usepackage{graphicx}
\usepackage{csquotes}
\usepackage{enumerate}
\usepackage{latexsym}
\usepackage{url}
\usepackage{subfig}
\usepackage{array}
\usepackage{spverbatim}
\usepackage{comment}
\usepackage{eurosym}
\usepackage{booktabs}
\usepackage{adjustbox}
\usepackage{subfig}
\usepackage{svg}
\usepackage{multirow}
\usepackage{adjustbox}
\usepackage{xcolor}

\begin{document}
\title{Experience and Prediction: A Metric of Hardness for a Novel Litmus Test}

\author{Nicos Isaak\inst{1}\orcidID{0000-0003-2353-2192}
	\and
	Loizos Michael\inst{1,2}}
\authorrunning{N. Isaak and L. Michael}

\institute{Open University of Cyprus\\
	\email{nicos.isaak@st.ouc.ac.cy} \ \ \ \ \email{loizos@ouc.ac.cy}
	\and
	Research Center on Interactive Media, Smart Systems, and Emerging Technologies}

\maketitle

\begin{abstract}
In the last decade, the Winograd Schema Challenge (WSC) has become a central aspect of the research community as a novel litmus test. Consequently, the WSC has spurred research interest because it can be seen as the means to understand human behavior. In this regard, the development of new techniques has made possible the usage of Winograd schemas in various fields, such as the design of novel forms of CAPTCHAs.

Work from the literature that established a baseline for human adult performance on the WSC has shown that not all schemas are the same, meaning that they could potentially be categorized according to their perceived hardness for humans. In this regard, this \textit{hardness-metric} could be used in future challenges or in the WSC CAPTCHA service to differentiate between Winograd schemas.

Recent work of ours has shown that this could be achieved via the design of an automated system that is able to output the hardness-indexes of Winograd schemas, albeit with limitations regarding the number of schemas it could be applied on. This paper adds to previous research by presenting a new system that is based on Machine Learning (ML), able to output the hardness of any Winograd schema faster and more accurately than any other previously used method. Our developed system, which works within two different approaches, namely the random forest and deep learning (LSTM-based), is ready to be used as an extension of any other system that aims to differentiate between Winograd schemas, according to their perceived hardness for humans. At the same time, along with our developed system we extend previous work by presenting the results of a large-scale experiment that shows how human performance varies across Winograd schemas.

\end{abstract}

\section{Introduction}
Since the late fifties, the AI community is concerned with endowing machines with commonsense and reasoning \cite{levesque2014our,kn:Valiant:2006knowledge}.  To that end, a number of challenges have been proposed to advance the field of AI, aiming to behoove AI researches build systems able to help or replace humans in day-to-day life. One of these challenges is the WSC, a carefully-crafted pronoun resolution task and a variant of the well known Recognizing-Textual-Entailment challenge (RTE) \cite{dagan2005pascal,michael2009reading} that is able to capture basic human abilities. 

Given a Winograd schema, people can anticipate and reason about causes and effects \cite{sap2019atomic}, and tell you who did what to whom, when, where, and why \cite{Gary_Marcus}. For instance, if someone tells us that, ``The city councilmen refused the demonstrators a permit because they feared violence."  and asks us ``who feared violence?", we can easily infer that the correct answer is the ``city councilmen". This example shows how humans, through commonsense and reasoning, can answer such questions. On the other hand, we know that current AI systems do not have that day to day commonsense and reasoning that humans do \cite{Gary_Marcus,levesque2014our}.

In a recent work, Bender \cite{kn:bender2015establishing} established a human baseline for the WSC, where it was shown that adults can tackle the challenge with a mean of 92\%. Along with the results, the authors have shown the importance of having humans to evaluate the schemas upon designing, as not all schemas have the same perceived hardness for humans \cite{kn:bender2015establishing,morgenstern2016planning}. Given that the WSC was developed to help humans design systems that mimic human behavior, it seems that shedding light on the perceived human hardness for schemas would be useful for the challenge itself. In this regard, this metric of hardness could be used, i) to categorize schemas according to the strengths and weaknesses of a particular group of participants, and, ii) in the WSC CAPTCHA service that uses Winograd schemas to identify humans from bots \cite{GCAI-2018:Using_Winograd_Schema_Challenge}.

In a past work of ours, we approached this problem by reusing the Wikisense system \cite{Isaak2016} for resolving Winograd schemas, and, roughly, by using the amount of training data it requires to correctly answer a given schema as an indicator of its hardness. This resulted in a system \cite{GCAI-2018:Data_Driven_Metric_of_Hardness} that correlates well with the performance of humans, albeit with limitations regarding the number of schemas it could be applied on and the time needed for the whole process, which was found to be very time consuming. To do that, we compared the Wikisense-approach results to humans' performance on a dataset of 143 schemas \cite{kn:bender2015establishing}.

In this work, we consider a new novel approach called WinoReg (from Winograd-Regression) which, through machine-learning, can deliver faster and more accurate results than our previous work. To that end, we build a new system that works within two different approaches, i) the Random-Forest approach, which directly relates with feature engineering, and, ii) the LSTM-based approach, which requires access to the hardness indexes of more Winograd schemas. In this regard, we extended Bender's work with a study that we designed and undertook, which involved 306 crowdsourced workers and 943 schemas. 

Within both approaches WinoReg proceeds by first training the regression model, and then using the learned model for faster computation during its deployment. Regarding the feature engineering of the Random-Forest approach, these features come from a number of works in the literature that have developed WSC-related systems, which we have re-implemented as needed \cite{kn:budukh2013intelligent,Isaak2016,kn:peng51solving,kn:Rahman:2012:RCC:2390948.2391032,kn:sharma2015towards}.

In the next sections, we start by presenting the challenge itself. We continue with our motivation section followed by the human-adult performance section. A high-level analysis of WinoReg's architecture is outlined in the fifth section, whereas a more detailed analysis of the Random Forest and the LSTM-based approach is given in the next two sections. We present the experiments along with our results in section eight. Finally, in the next sections, we present some highlights of previous work along with potential implications and recommendations for future research.

\section{Challenge Basics}
Broadly speaking, the WSC is about resolving ambiguities because the information needed is not grammatically present in the examined schemas. Consequently, each Winograd schema comprises two halves, with each half consisting of a sentence, a definite pronoun or a question, two possible pronoun targets (answers), and the correct pronoun target \cite{levesque2012winograd}. The following schema (a pair of halves) illustrates the key characteristics of the challenge:
\subsubitem First-half: \textit{Sentence: The city councilmen refused the demonstrators a permit because they feared violence. Question: Who feared violence? Answers: The city councilmen, The demonstrators. Correct Answer: The city councilmen.} 
\subsubitem Second-half: \textit{Sentence: The city councilmen refused the demonstrators a permit because they advocated violence. Question: Who advocated violence?  Answers: The city councilmen, The demonstrators. Correct Answer: The demonstrators}. 

Given just one of the halves, the aim is to resolve the definite pronoun through the question to one of its two co-referents. To avoid trivializing the task, the co-referents are of the same gender, and both are either singular or plural. Moreover, the two halves differ in a special word or phrase that critically determines the correct answer. Schemas that do not \textit{strictly} follow these rules are called ``schemas in the broad sense''. It is believed that the WSC can provide a more meaningful measure of machine intelligence when compared to the Turing Test \cite{levesque2014our}. Its a challenge that has been proposed as the means to understand human behaviour \cite{levesque2014our}. In this sense, statistical resolvers would not be able to accumulate tricks or discover patterns of words to tackle it \cite{levesque2014our}. This might happen because of the presumed necessity of reasoning with commonsense knowledge to identify how the special word or phrase affects the definite pronoun's resolution. The challenge is already in full swing with other AI challenges that aim to tackle the goal of endowing machines with human commonsense and reasoning. By extension, it is believed that a system that contains the commonsense knowledge to resolve Winograd schemas correctly should be capable of supporting a wide range of AI applications \cite{levesque2012winograd}.

The WSC has been a topic of interest for several years. Among the steps taken towards developing systems able to tackle the challenge has been the development of relevant datasets. According to the literature, multiple well-known datasets exist, like: 1) The original collection of Winograd schemas referred to as WSC273, WSC285, WSC286, or WSC288 \cite{levesque2012winograd};
2) The Definite Pronoun Resolution (DPR) dataset, a variation of the Winograd Schema Challenge developed by Rahman \& Ng \cite{kn:Rahman:2012:RCC:2390948.2391032};
3) The Pronoun Disambiguation Problem (PDP) dataset, which was collected from the literature to be used as a testing set for the first Winograd Schema Challenge, which took place in 2016 \cite{morgenstern2016planning};
4) The WinoGrande dataset \cite{sakaguchi2020winogrande}, which is a large-scale dataset collected via crowdsourcing on Amazon Mechanical Turk;
5) The WinoFlexi dataset, which is similar to the original WSC dataset, collected via crowdsourcing on the MicroWorkers platform \cite{10.1007/978-3-030-35288-2_24};
6) The Winograd Natural Language Inference dataset that is a part of the GLUE benchmark \cite{wang2019glue};
7) Other datasets that were developed to measure gender bias among Winograd schemas \cite{rudinger-etal-2018-gender,zhao-etal-2018-gender}.

As stated by Kocijan et al. \cite{kocijan2020review}, three different approaches have been used to tackle the challenge: i) feature-based approaches that basically try to tackle it by extracting semantic information from several sources,  ii) neural-based approaches that are trained on unstructured or pre-trained data, and iii) language-model approaches that utilize large-scale pre-trained language models and are sometimes fine-tuned on a specific WSC dataset to maximize their performance. At the time of writing, we can say that the challenge can be tackled with an average score of 70\% although there are various approaches able to solve only a subset of schemas \cite{kocijan2020review}. For instance, systems are able to tackle the original dataset (WSC285) with an average score of 66 \% (lowest 42\% and largest of 90\%), the Definite Pronoun Resolution (DPR) dataset with an average of 78\% (lowest 63\% and largest of 93\%), and the PDP dataset can be tackled with an average score of 77\% (lowest 58\% and largest of 90\%). A more deep analysis about the datasets and systems able to tackle the challenge can be found in \cite{kocijan2020review}.

\section{Motivation}
It is widely believed that well-constructed Winograd schemas are easy for humans and hard for machines because they require the use of commonsense knowledge to correctly resolve the definite pronoun \cite{levesque2014our}. According to Levesque, in every schema, you need to have background knowledge that is not revealed in the words of the sentence to be able to clarify what is going on \cite{levesque2012winograd}. 

Broadly speaking, due to schema discrepancies, not all Winograd schemas are equally easy or hard for humans, and the task of being able to predict their hardness index is an interesting question. Additionally, with every single schema any potentially developed system should presumably be able to demonstrate how humans tackle it, meaning, that, there are different kinds of schemas. 

What we know about the perceived human hardness index on the WSC is largely based on Bender’s work \cite{kn:bender2015establishing}, who, through an experiment he undertook, identified that human adults tackle the WSC with a mean of 92\%. In a past work of ours  \cite{GCAI-2018:Data_Driven_Metric_of_Hardness}, towards answering the previous question, we started by considering the Wikisense system \cite{Isaak2016}, which is a commonsense and reasoning system able to resolve a number of Winograd schemas. Basically, Wikisense parses each examined schema to identify the necessary keywords to search for relevant Wikipedia sentences. Next, for every Wikipedia sentence, it returns semantic scenes, which are triples based on nominal-subjects and direct-objects returned by a dependency parser. These semantic scenes are fed to a Learner that constructs the necessary knowledge, which can be searched through a Reasoner for the tackle of the challenge. 

Given that Wikisense gets its training data in real time from the English Wikipedia, we developed a new system ---Wikisense-based approach--- whose performance improves as it gets more training data while its trying to resolve a given schema. Specifically, we have found that the amount of training data needed for the resolving of a given schema correlates positively with the perceived human hardness index of that specific schema. However, the resulting model was able to offer the hardness index on only 57\% of our tested schema halves, which is in direct relation with the keyword implementation of Wikisense that is based on the semantic analysis of the given schemas; If Wikisense cannot extract a keyword then the Wikisense-based approach cannot return the hardness index of the examined schema. Additionally, because of its dependency on training during query-answering, it was found that the Wikisense-based system needs, on average, eight hours to output the hardness index of given schema half. 

There are systems that are already in full-swing with the Wikisense-based approach, meaning that they already use its mechanisms to differentiate schemas according to their perceived hardness for humans. In this regard, WinoFlexi \cite{10.1007/978-3-030-35288-2_24}, which is a crowdsourced collaboration platform for the development of Winograd schemas from scratch, leverages the Wikisense-based approach to provide feedback to workers regarding the \textit{quality} of their developed schemas. In this regard, we can say that the Wikisense-based approach results are disproportional to the demand of new developed schemas. Additionally, in an earlier work of ours we have demonstrated how the WSC can form a novel form of CAPTCHAs \cite{GCAI-2018:Using_Winograd_Schema_Challenge}, with the ultimate goal of bringing more AI researchers to work on the challenge. Like in every other CAPTCHA service, there is a high demand of new Winograd schemas which could serve as the means to identify fraudulent actions. In this regard, systems like the Wikisense-based approach could be used to make sure that the CAPTCHA service would display harder schemas to solve in the case of possible fraudulent actions. Furthermore, in the case of humans it could be used to ensure that the generated instances are not overly demanding.  

As stated in the literature \cite{kocijan2020review}, neural approaches, and specifically, language-model-based approaches that were trained on a large corpus of text were able to tackle the challenge with 90.1\% accuracy on WSC273 \cite{sakaguchi2020winogrande}, which is the highest performance achieved on the original dataset by a large margin. On the other hand, this does not mean that the challenge itself is tackled or that language models are able to show commonsense-reasoning abilities like humans do. 
For instance, the model by Sakaguchi et al. \cite{sakaguchi2020winogrande}, might have performed well because, through training, it was able to exploit a systemic bias within the dataset that helped it tackle the challenge with high accuracy. Specifically, this might have happened because language models or just neural networks that just predict probabilities of the next or previous word in a sentence may have their limits \cite{brown2020language}, meaning that more text does not always yield better results. In general, heuristics or methods non-crucial to the WSC-idea, such as the choice of word embeddings or language models can easily affect the results \cite{kocijan2020review}. It seems that the problem that Levesque tried to avoid by introducing the challenge as an alternative to the Turing test still remains: systems are able to discover tricks or systematic bias in words to tackle the schemas without being able to show commonsense and reasoning abilities. For instance, recent experiments have shown that state-of-the-art language models that are currently being able to tackle challenges like the WSC struggle to tackle challenges that directly relate to abductive-reasoning, meaning they lack reasoning abilities which are trivial for humans \cite{bhagavatula2019abductive}.

Having systems able to tackle schemas close to human performance led researchers to investigate how far they could push the difficulty level of Winograd schemas, proposing at the same time various mechanisms to build challenging ones \cite{eniac}. It was stressed out by the creators of the challenge that clever tricks involving features or groups of words should be avoided or eliminated. Of course, recent language-model achievements on the challenge itself indirectly show that there are still biases or hidden patterns that are easily discoverable by these pattern-extracting solvers, meaning that the ability to tackle a variety of schemas does not show commonsense and reasoning abilities like humans do \cite{eniac,kocijan2020review}. As stated by Cozman et. al. \cite{eniac} the solution to this problem is the development of harder schemas, where no word patterns or rooted social-norms can be tracked and cracked by language learners without deep understanding. However, this should be done with respect to the effort needed to solve a given schema by both humans and machines, although evaluating the degree of common sense possessed by a machine remains difficult \cite{talmor2018commonsenseqa}.

To the best of our knowledge, the Wikisense-based approach is the only system that can differentiate between Winograd instances, albeit with various limitations that raise questions regarding whether we should look for alternative solutions based on different techniques. In this regard, to find a faster and more accurate way to output the hardness index of Winograd schemas, we consider WinoReg, which is a system based on a machine-learning approach. Through experience and prediction, WinoReg learns how to compute the hardness of a given schema based on two different approaches, i) the Random-Forest, and ii) the LSTM-based approach. Before proceeding with WinoReg's architecture, it is interesting to briefly review Bender's work regarding human perceived hardness on the Winograd schemas.

\section{Human-adult Performance on the WSC}
Bender \cite{kn:bender2015establishing}, through an experiment he undertook, which involved the participation of adult English speakers, identified that human adults tackle the WSC with a mean accuracy of 92\%. Furthermore, it was found that adults need, on average, 15 seconds to answer a given schema.

To the best of our knowledge, this is the only set available to provide us with the necessary training and testing data \cite{GCAI-2018:Data_Driven_Metric_of_Hardness}. In his work, he used schemas developed by experts ---called as the original dataset \cite{levesque2012winograd,morgenstern2016planning}--- which, at the time of writing, consisted of 143 schemas (286 schema halves). The experiment ran on Amazon's Mechanical Turk where 407 adult speakers, who speak English fluently, participated. Results showed that adult speakers are, on average, able to correctly resolve 92\% of the Winograd schemas, which sets the bar very high, compared to what systems can achieve \cite{kocijan2019surprisingly,liu2016probabilistic,kn:peng51solving,kn:Rahman:2012:RCC:2390948.2391032}. On the other hand, in the experiments it was shown that there are schemas that are harder to resolve than others; for instance there are schemas that humans scored a mean of 45\%. A detailed analysis of human performance on each individual WSC instance (accuracy) is available from: \url{https://github.com/benderdave/wsc-exp.git}.

\section{The High-Level Architecture of WinoReg}
\begin{figure}[h]
	\centerline{\includegraphics[width=\columnwidth]{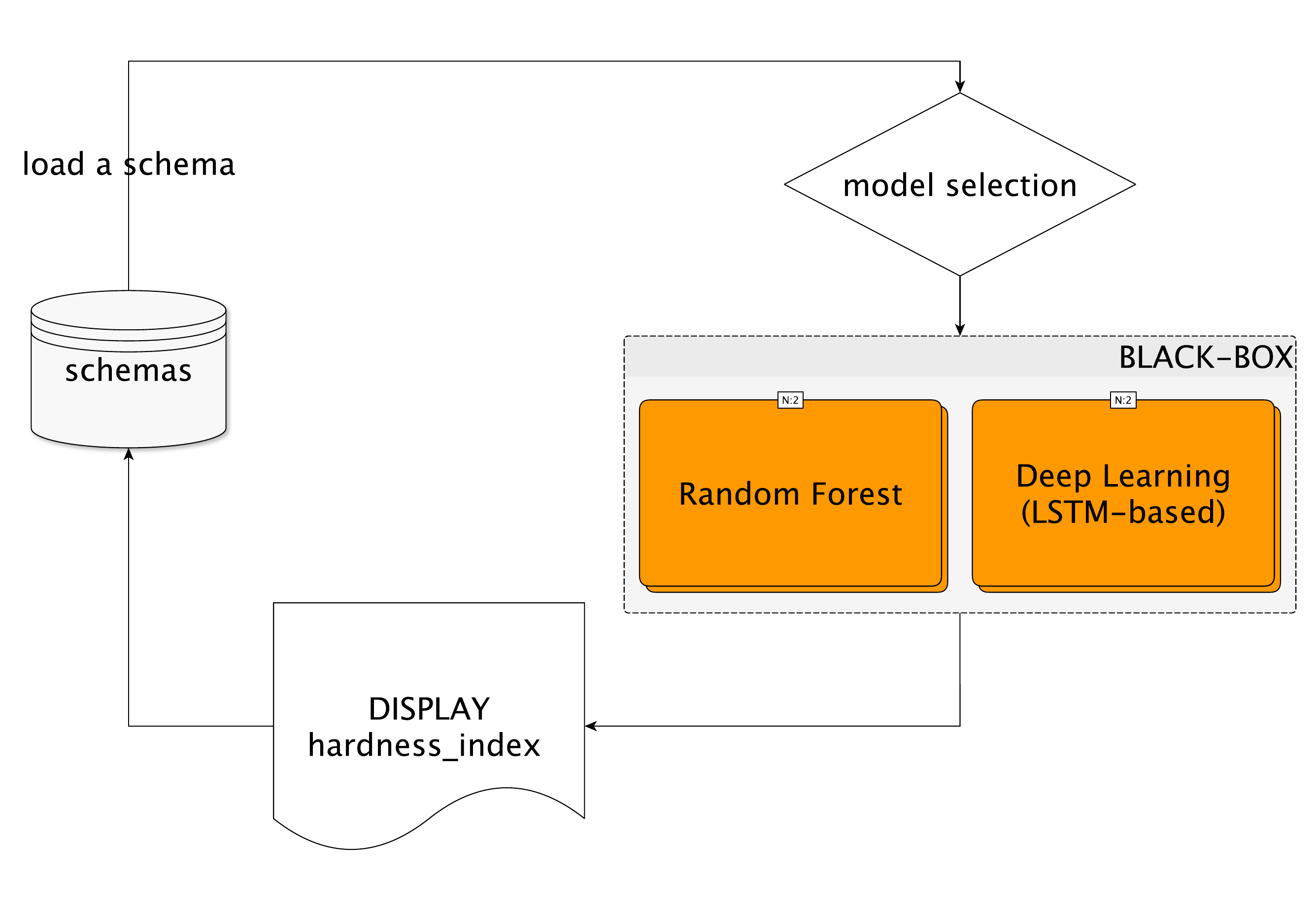}} 
	\caption{WinoReg's Architecture to compute the hardness indexes of Winograd schemas. The black-box shows that the system can work in two distinct modes.}
	\label{fig:mainwinoreg}
\end{figure}

Here, we present the high-level architecture of WinoReg (see Figure \ref{fig:mainwinoreg}). WinoReg works in two operational modes, namely, the random-forest, and the LSTM-based mode (deep learning). In both modes, it outputs the hardness of any schema through regression analysis, where it examines the relationship between the schema halves and the perceived human hardness indexes \cite{kn:bender2015establishing}. 

After the training it uses the learned model for faster computation during its deployment. Regarding the feature engineering of the Random-Forest approach, these features come from a number of works in the literature that have developed WSC-related systems, which we have re-implemented as needed \cite{kn:budukh2013intelligent,Isaak2016,kn:peng51solving,kn:Rahman:2012:RCC:2390948.2391032,kn:sharma2015towards}. Specifically, within the Random Forest mode, WinoReg analyzes each schema to output a required number of features. Next, all of the features are given as an input to the learned model to output the hardness of a schema half. On the other hand, within the LSTM-based approach, WinoReg does not require to estimate the values of features, meaning that any given schema can be given directly to the model to acquire its hardness index. In both cases, WinoReg can load a schema from a schema-database to output its hardness index, which is a value in the range of 0-1. Compared to the Wikisense-based approach, no schema is discarded.

\hfill\\
In the next sections, we will show how WinoReg works, based on the approaches above. Specifically, in the first part, we will discuss how the engine estimates the values of features to build the Random Forest model, and, in the second part, we will show how deep learning comes into play.

\section{WinoReg: A Random Forest Approach}
\begin{figure}
	\centerline{\includegraphics[width=\columnwidth]{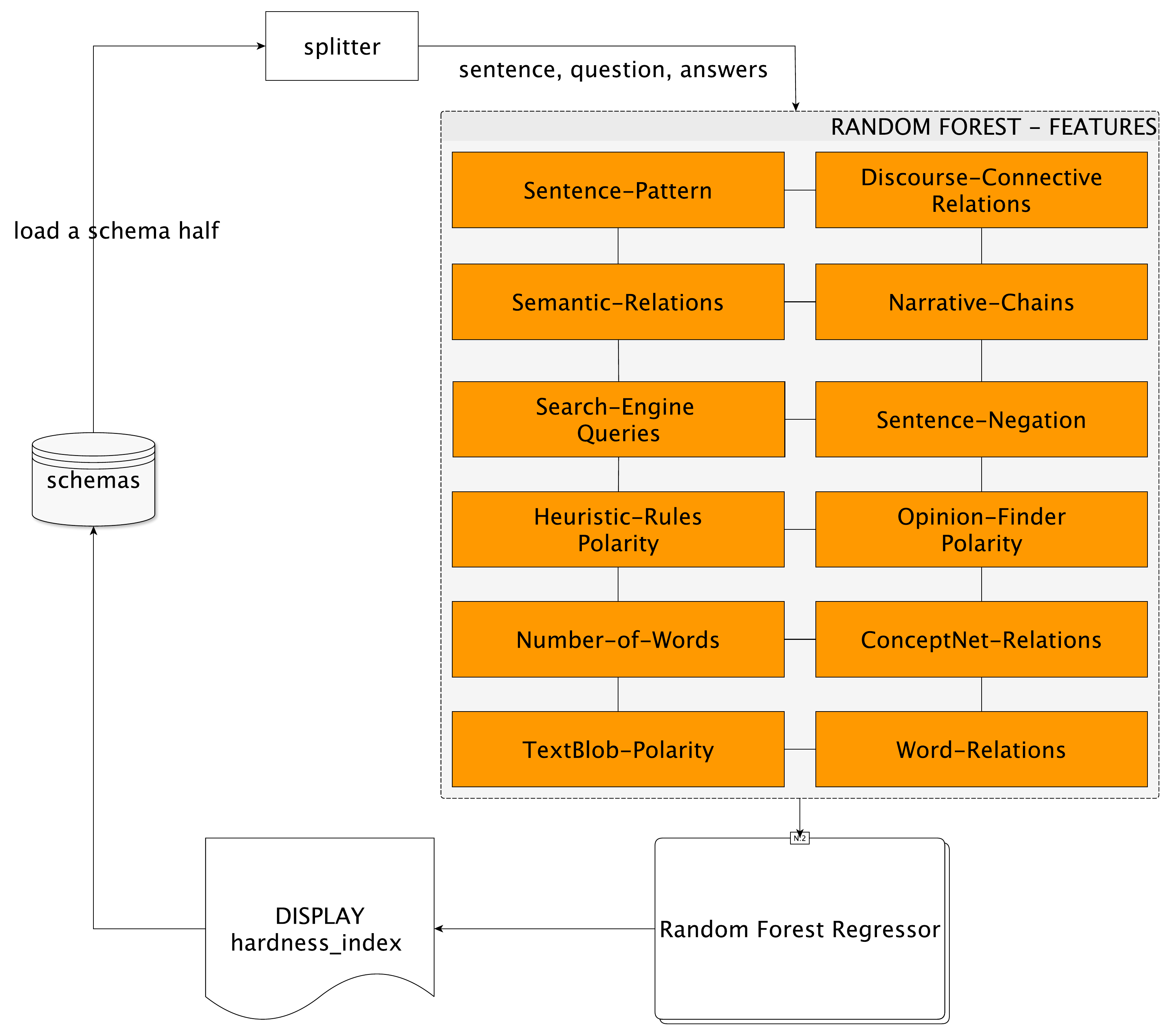}} 
	\caption{WinoReg's Architecture based on Random Forest: Given a Winograd schema WinoReg outputs the perceived human hardness index.}
	\label{fig:randomforesthardness}
\end{figure}

Within this approach, WinoReg is based on training a regression model with the use of Decision Trees. We use, in particular, the Random Forest algorithm \cite{fry2018hello}, which was introduced in 2001 \cite{breiman2001random}. The Random Forest algorithm, which involves the construction of an ensemble of Decision Trees, each trained on random subsets of the data, showed significant improvements in accuracy of different kinds of problems \cite{breiman2001random}. A recent line of research showed that it is one of the best algorithms that maintain high imputation performance on linear regression across a range of performance metrics \cite{10.1007/978-3-030-35288-2_18}. Like any other Machine Learning algorithm, the focus of the Random Forest algorithm is to form a rule with reasonable accuracy, which could be used as a prediction tool on future data \cite{probst2019hyperparameters}. In this regard, we aim to train a model using the Random Forest algorithm able to estimate the perceived human hardness index of Winograd schemas (see Figure \ref{fig:randomforesthardness}).

\subsection{Feature Preparation }
Within ML, we need to transform our data to find the appropriate representations to make it more manageable to the task at hand \cite{franccois2017deep}. As we want to estimate the hardness index of a schema half, which indirectly relates to the selection of the correct answer, our Random-Forest approach expects features related to the schema half parts ---sentence, question, and the two pronoun targets (candidates). Compared to Wikisense-based system, WinoReg does not make use of the correct answer of each schema. 

To train our system we use 50 features from 12 components that we built from scratch. The majority of these features are based on non open-source systems, from the literature, that were previously used to tackle the challenge \cite{kn:budukh2013intelligent,GCAI-2018:Data_Driven_Metric_of_Hardness,kn:peng51solving,kn:Rahman:2012:RCC:2390948.2391032}. Most of these features relate to semantic relations that are taken from each examined schema. In this regard, our system uses the \textit{spaCy}\footnote{\url{https://spacy.io}\\spaCy's statistical model: en\_core\_web\_sm} dependency parser to turn raw-text into semantic relations; these are relations that show how the sentence words are related to each other. According to the literature, the semantic relations are considered good if they can express the structure of the text and can differentiate, at the same time, between the events and their participants \cite{kn:sharma2015towards}. In this regard, via spaCy, we can output relations that show how the pronoun targets relate to the definite pronoun and the events in which they participate \cite{Isaak2016,icaart20}. 

For instance, consider the following schema half (referred to later as the \textit{catch} example): \textit{Sentence: The cat caught the mouse because it was clever. Question: Who is clever? Answers: The cat, The mouse}. Via spaCy, we can output three semantic relations, which tell us that  ``a cat caught a mouse", and ``something/someone is clever":
\subsubitem [cat-noun, caught-verb, mouse-noun]
\subsubitem [it-pronoun, was-aux-verb, clever-adj]
\subsubitem [was-aux-verb, caught-verb]


A detailed analysis of the feature development process is given in the next paragraphs.

\subsection{Sentence-Pattern}
In a recent work, we have shown that the structure of each schema-half's sentence plays an important role in its quality \cite{10.1007/978-3-030-35288-2_24}. It seems that schemas that are developed using a variety of sentence patterns/types ---complex, compound-complex---, are preferable than schemas that are based on simple types. In this regard, to design our first feature we make use of a tool that is able to output the sentence-type of each examined schema \cite{10.1007/978-3-030-35288-2_24} (stored as ST). Given any English sentence, the tool is able to output its type which can be either a simple, a compound, a complex, or a compound-complex sentence. At the same time, it outputs its pattern/clause (e.g., ``SV because SV'', ``SV and SV because SV.'', ``Cause/Effect"), which directly relates to the connectors that each sentence uses between its clauses (stored as SP). Hence, within this component we are able to engineer two features, namely ST and SP.

\subsection{Sentence-Negation}
It is widely accepted that \textit{negation} plays an important role in capturing the semantics of text, as it is used to reverse the polarity of parts of a statement \cite{blanco2011some,Isaak2016}. To encompass these kind of rules, we analyze each schema half to estimate if the two candidates and the definite pronoun are governed by negation; this is done via the sentence and question triples of the schema half (see the \textit{catch} example). In this regard, from the \textit{negation}-component we create two binary features, (\textit{STN} for the two candidates, and \textit{QTN} for the definite pronoun) that contain the value of 1, if negation exists, and, otherwise, the value of 0.

\subsection{Schema's Semantic-Relations}
This component directly relates to the semantic relations of a given text. As stated in the literature \cite{kn:Rahman:2012:RCC:2390948.2391032}, via web queries we might phase precision and recall problems. Specifically, when a pronoun target and a verb appears next to each other, it does not mean that a subject-verb relation exists between them (a problem of precision). On the other hand, these queries fail to obtain subject-verb relations where a pronoun target and verb are not close to each other (a problem of recall). To eliminate these kinds of problems, we search Wikisense's Wikipedia-corpus to see how many times each pronoun target appears as subject or as an object. If the definite pronoun appears as a subject in a triple relation, we search to find which pronoun target appears as a subject most of the times; Otherwise, if the pronoun appears as an object, we search to find which pronoun target appears most of the times as an object. From the semantic-relation component we create a single feature SEM, which equals 1 if the definite pronoun has the same role as the first pronoun target, otherwise 2 if it has the same role with the second pronoun target. If we cannot determine their roles then SEM equals -1.

\subsection{Number-of-Words}
It seems that the sentence length of each schema directly relates to the resolution of the definite pronoun \cite{icaart20}, where schemas with longer sentences tend to be harder to answer. In accordance with our findings, we engineered a feature that directly relates, in terms of words, to the length of each sentence (SL). 

\subsection{Word-Relations}
Word-relations features relate to candidate-independent, and candidate-dependent relations, where, according to previous works \cite{kn:Rahman:2012:RCC:2390948.2391032}, they seem to play an important in the tackle of the WSC. The only catch is that they can only be applied in sentences that contain a connective (Cn) word (e.g., because). In this regard, for the candidate-independent features we create two features (WN, WP), where, WN refers to the number of words in each sentence (except the two candidates and the Cn), and, WP refers to the number of word pairs; these are pairs of words appearing before Cn with each word appearing after Cn, excluding adjective-noun pairs, noun-adjective pairs, and the two candidates. 

For the candidate-dependent features we engineer three features, namely HN, VF, and AF. Specifically, HN contains the number of the head words of the two candidates that were returned by the dependency parser; if we cannot determine the two candidates in the sentence then the HN feature is set to 0. Subsequently, the VF feature contains the number of the verbs, and JF the number of the adjectives that modify the two candidates.

\subsection{Search-Engine Queries}
Recent work has shown that search-engine queries are able to provide us with world knowledge, which is useful for the tackle of the challenge \cite{kn:peng51solving,kn:Rahman:2012:RCC:2390948.2391032,kn:sharma2015towards}.
For instance, in the \textit{catch} example, we can acquire world knowledge to learn that someone who is clever can easily catch other things, which leads us to resolve the definite pronoun to the \textit{cat}. In this regard, as other works have shown, we follow a similar approach to build features that are based on search queries. 

For every schema we build six queries, namely QR1: A1VQ, QR2: A2VQ, QR3: A1VQW, QR4: A2VQW, QR5: JA1, QR6: JA2; A1 and A2 are the two candidates, VQ the question verb that governs the definite pronoun, W the sequence of words following VQ in the question, and J the question adjective that follows a verb-to-be. For instance, for the \textit{catch} example we generate and search the Google search-engine with the following queries: (QR1) “cat was”; (QR2) “mouse was”; (QR3) “cat was clever”; (QR4) “mouse was clever”; (QR5) “clever cat”; and (QR6) “clever mouse”. Next, using the number of hits that were returned by the search engine, we built eight binary features ---GL1i1, GL1i2, GL2i1, GL2i2, GL3i1, GL3i2, GL4i1, GL4i2---, as in Rahman \& Ng \cite{kn:Rahman:2012:RCC:2390948.2391032}. The first two features (GL1i1, GL1i2) are computed from QR1 and QR2, the next two (GL2i1, GL2i2) from QR3 and QR4, and the third (GL3i1, GL3i2) from QR5 and QR6 (The last two features are computed based on the results returned from all of the queries). For instance, if the absolute value of  |QR1, QR2|, is bigger than the threshold of 20\% ($th$) in favor of the first candidate, then, GL1i1 equals 1 and GL1i2 equals 0; if the opposite exists then GL1i1 is set to 0 and GL1i2 to 1. To estimate the other features we follow a similar approach; more details about the procedure can be found in the paper where it was originally introduced \cite{kn:Rahman:2012:RCC:2390948.2391032}.

Recent experiments with GPT3 language-model \cite{brown2020language} have shown potential contamination in their training set while tackling the WSC or other similar tasks. This relates with text found in the WWW which contains WSC schemas or similar discussions that might help relevant models or a specific search engines find cues they were not supposed to find. Although this is a challenging task that needs to be examined further when designing benchmarks and when training models \cite{brown2020language}, both the way the search queries are constructed and the defined threshold of 20\% help avoid problems that relate with potential contamination.

To avoid problems with proper-names (persons) where we cannot retrieve search query hints we make use of Framenet \cite{baker1998berkeley}. As stated in other works \cite{kn:budukh2013intelligent,kn:Rahman:2012:RCC:2390948.2391032}, it is unlikely that search engines will return meaningful counts for persons. In this regard, in a schema where the candidates are proper names we search Framenet to find and substitute them with their roles. Specifically, for every triple relation we search Framenet for NP.EXT and NP.OBJ relations, where, NP.EXT shows the subjects and NP.OBJ the objects of the corresponding event (for instance, in the \textit{catch}, if instead of a cat and mouse we had persons then we would search Framenet for the event catch). In case of successful search from Framenet we replace the persons with their Framenet roles. Consequently, we form six queries and search the Google engine to generate eight features: GLF1i1, GLF1i2, GLF2i1, GLF2i2, GLF3i1, GLF3i2, GLF4i1, GLF4i2.

\subsection{ConceptNet-Relations}
ConceptNet is a freely available semantic commonsense toolkit \cite{liu2004conceptnet}. Its knowledge-base is a semantic network, where nodes are the concepts and edges the relations among them. It is like a parser that describes and expresses general human knowledge from sentences that were automatically acquired from the Open-mind Common-Sense project \cite{liu2004conceptnet,singh2002open,speer2017conceptnet}. It contains concepts about common basic knowledge about various facts, connected with other facts, using different kind of relations (e.g., \textit{relatedTo, AtLocation, IsA, PartOf}) \cite{kn:budukh2013intelligent}. Our system, makes use of ConceptNet to find possible relations between the two candidates and the word ---verb, adjective--- that governs the definite pronoun; this is done by a ConceptNet function that returns a value in the range of 0-1, where, the higher the value the higher the relatedness is. In this regard, we engineer a feature, (called CN) that equals 1 if the relatedness value of the first candidate is greater then the value of the second candidate; if the opposite exists then then value of CN equals 2, and, if we cannot find any difference, it equals -1. Additionally, like before, we consider Framenet \cite{baker1998berkeley} for issues with proper names, and create the CNF feature, where its values are being computed in the same way as the CN values.

\subsection{Discourse-Connective Relations}
As reported by Rahman \& Ng \cite{kn:Rahman:2012:RCC:2390948.2391032}, causal relations, which are signaled by discourse connectives, show the world knowledge between events. For instance, in the sentence, ``The lion eat the zebra because it was hungry", there is a causal relation, which is given by the discourse connective ``because", between the events ``eat" and ``hungry"; this \textit{causal} relation help us resolve the definite pronoun ``it" to the lion.

For each schema half, we search the Wikisense corpus for a triple of the form (V, Cn, X), and count its frequencies of occurrence; Cn is a discourse connective, V is a verb in the clause that governs the two candidates, and X is a stemmed verb or an adjective that governs the definite pronoun. Each triple has to be validated through the following procedure: i) we search the Wikipedia corpus to find its frequencies of occurrence; ii)  if the the number of occurrences is at least 100 then we proceed to the next step \cite{kn:Rahman:2012:RCC:2390948.2391032}; iii) if X is a verb, then it resolves the pronoun to the candidate that has the same role as the definite pronoun; otherwise, if the sentence does not involve comparison and X is an adjective, it resolves the pronoun to the candidate that serves as the subject of V. To encode this heuristic decision we create a binary feature (CNT); CNT equals 1 if the definite pronoun is resolved to the first candidate, and 2 if it is resolved to the second candidate. Otherwise, in case we cannot resolve the definite pronoun, CNT equals -1.

\subsection{Event-Chaining via Narrative Chains}
\textit{Narrative-chains} are sequences of events, in a story that shows the role of the protagonist/actor, which is denoted as \textit{-s}: subject or \textit{-o}: object \cite{kn:budukh2013intelligent,kn:Rahman:2012:RCC:2390948.2391032}. To the best of our knowledge, one of the best available narrative-chain datasets is the Chambers and Jurafsky's narrative chains \cite{kn:chambers2008unsupervised}; these are ordered sets of 12 events (verbs) centered around a common protagonist that show its role in the chain (subject or object).

For every schema half, we determine the events the two candidates and the definite pronoun participate in along with their protagonist role (subject or object). For instance, in the next schema half: \textit{Sentence: The city councilmen refused the demonstrators a permit because they advocated violence. Question: Who advocated violence? Answers: The city councilmen, The demonstrators.}, via Wikisense mechanisms we output two triples: i) refused (x-subject, y-object), ii) advocate (they-subject, violence-object). Hence, in this example, we want to determine the protagonist of the refuse-? event, that participates in the advocate event as a subject (the definite pronoun ---they--- indicates the subject position).

Next, from Chambers and Jurafsky’s, and for each such pair, we extract all the chains that contain both elements (refuse and advocate). For instance, in our example, Chambers and Jurafsky narrative chain contains \textit{refuse-o and advocate-s}, meaning that the protagonist in this chain is the object of a refuse event and the subject of an advocate event (the demonstrators); If WinoReg cannot find narrative chains containing both elements, it runs again the same procedure but with a similarity mechanism enabled. In the end, we create a feature (NCH) that equals 1 if the answer is the first candidate, and 2 if it is the second candidate. Otherwise, if we cannot output triples or find any narrative chains, it equals -1.

\subsection{Event-Polarity with Heuristic Rules}
Word polarity, which has been widely studied in the NLP field \cite{hassan2010identifying}, can help us to resolve the definite pronoun in specific schema halves \cite{kn:budukh2013intelligent,kn:peng51solving,kn:Rahman:2012:RCC:2390948.2391032}. This is a straightforward procedure that can be summarized in three steps: i) find the polarity of the definite pronoun; ii) determine the polarity of the two candidates; iii) select the candidate that has the same polarity as the definite pronoun.
To find the polarity values we use the Wilson et al. subjectivity lexicon \cite{wilson2005recognizing}, a lexicon that assigns to various events their polarity, such as negative, positive, or neutral.

Let us use the following example to explain the procedure we follow to assign the polarity values: \textit{Sentence: The city councilmen refused the demonstrators a permit because they advocated violence . Question: Who advocated violence?, Answers: The city councilmen, The demonstrators.} According to the schema half we know the following:
\subsubitem \textit{city-councilmen} is the subject of the event \textit{refuse}
\subsubitem \textit{demonstrators} is the object of the event \textit{refuse}
\subsubitem \textit{they} is the subject of the event \textit{advocate}.

From the Wilson et al. subjectivity lexicon \cite{wilson2005recognizing}, we acquire the polarity of the \textit{refuse} event, which is negative. In this regard, the polarity of the deep subject \textit{city councilmen} becomes negative and the polarity of the object \textit{demonstrators} becomes positive. Additionally, we know that the polarity of the event \textit{advocate} in the subjectivity lexicon is positive, hence the polarity of the definite pronoun \textit{they}, which participates in the subject of the event \textit{advocate}, becomes positive. Consequently, we can conclude that the polarity of both the definite pronoun and the \textit{demonstrators} is the same, which lead us to resolve the definite pronoun ---they--- to \textit{demonstrators}.  

The event-polarity procedure lead us to the engineering of six binary features, namely, RP1i1, RP1i2, RP2i1, RP2i2, RP3i1, RP3i2. Initially, all of these features are set to zero. The first two features, RP1i1, RP1i2 refer to the correct pronoun target, where, in our example are set to RP1i1=0 and RP1i2=1 (since the correct pronoun target ---demonstrators--- is the second one). The two other features (RP2i1 and RP2i2) are the concatenation of the polarity values, determined for both the definite pronoun and the two candidates; in our example, RP2i1=negative-positive, and RP2i2=positive-positive. 

To estimate RP3i1 and RP3i2, we simply take the previous features of RP2i1 and RP1i2 and append, if exists, the polarity reversing connective, such as \textit{although}, which is a connective that flips the polarity \cite{kn:Rahman:2012:RCC:2390948.2391032}.
Specifically, If a polarity reversing connective exists we simply take RP2i1 and RP2I2 and append the connective. For instance, RP3i1 = RP2i1 + connective, RP3i2 = RP2i2 + connective. Furthermore, we enhance the polarity features by creating an additional feature (RPTL) that shows the best pronoun target.To that end, we simply take the first two binary features (RP1i1, RP1i2), and generate a new one (RPTL). If RP1i1 $>$ RP1i2 then the value of RPTL equals 1, and, otherwise, if the opposite exists, the value of RPTL equals 2. If we cannot determine RP1i1 and RP1i2 then RPTL is set to -1.

\subsection{Event-Polarity with OpinionFinder}
This is a machine-based polarity that uses a sentiment-analyzer to resolve the definite pronoun of a schema-half. 
As stated in other works \cite{kn:peng51solving,kn:Rahman:2012:RCC:2390948.2391032}, instead of using a heuristic approach to estimate the polarity values, we use OpinionFinder \cite{wilson2005opinionfinder}, which is a machine driven approach able to perform subjectivity analysis. With tools like OpinionFinder we can easily annotate phrases with their contextual polarity values. To that end, we compute the OpinionFinder polarity features in the same way we did with the rule-based polarity features, and create seven features (OP1i1, OP1i2, OP2i1, OP2i2, OP3i1, OP3i2, OPTL).

\subsection{Event-Polarity with TextBlob}
Given that our previous polarity features are based on similar approaches, namely, Wilson et al. subjectivity lexicon \cite{wilson2005recognizing} and Wilson et al. OpinionFinder \cite{wilson2005opinionfinder}, here, we use another, simpler polarity mechanism ---called TextBlob-Polarity\footnote{\url{https://textblob.readthedocs.io/en/dev/}}. This is an NLP library that can process textual data and output, among others, the events' polarity values. Specifically, with the TextBlob's sentiment analysis we return the polarity of the verb that governs the two candidates, and the polarity of the verb that governs the definite pronoun. Finally, we create two features (TBSPOL, TBQPOL) that can be either neutral, positive, or negative.

\section{WinoReg: A Deep-Learning Approach}
\begin{figure}
	\centerline{\includegraphics[width=\columnwidth]{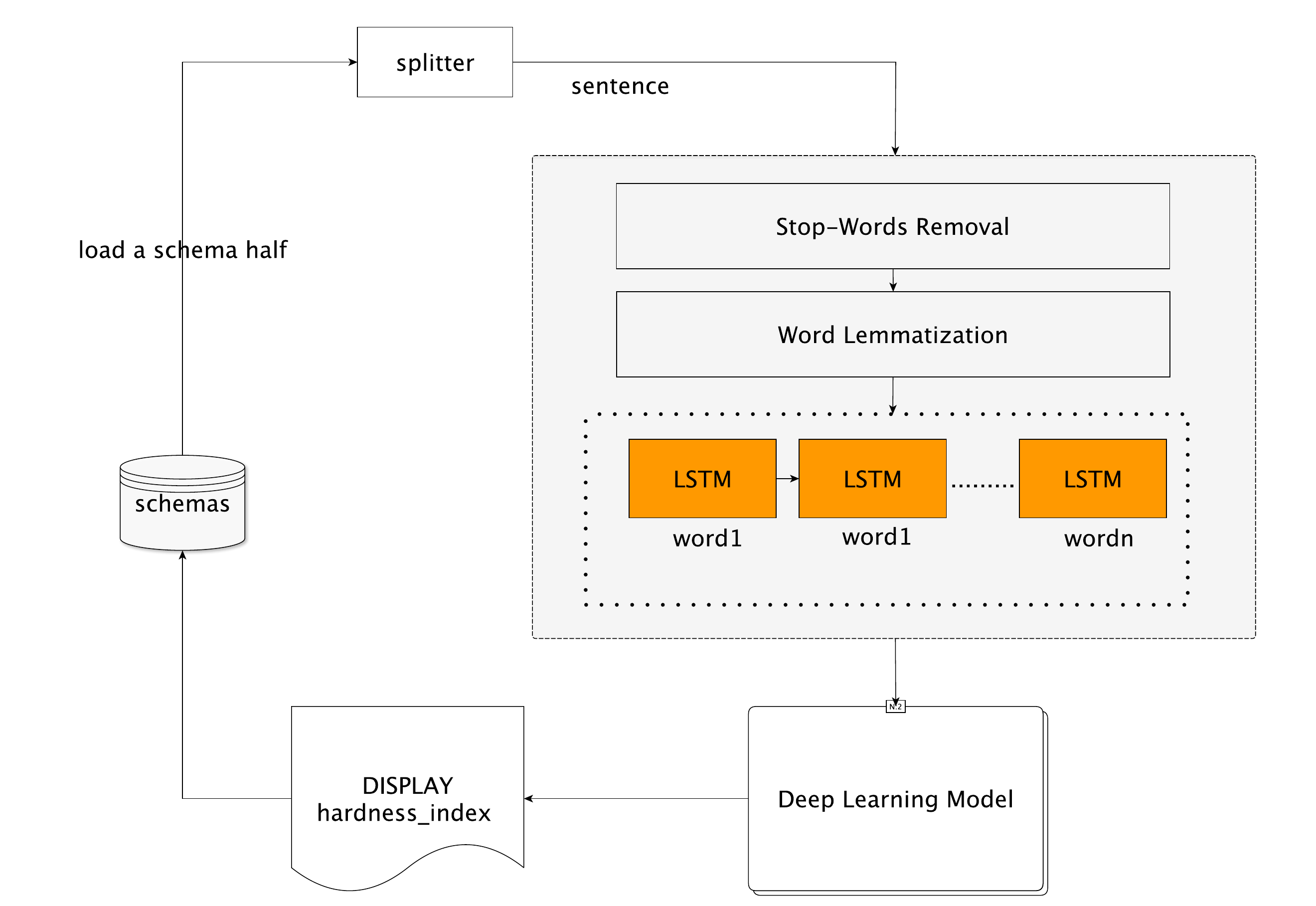}} 
	\caption{WinoReg's Architecture based on Deep-Learning: Given a Winograd schema WinoReg outputs the perceived human hardness index.}
	\label{fig:winoRegDL}
\end{figure}

Within this approach we train WinoReg using deep learning (see Figure \ref{fig:winoRegDL}), which is another increasingly popular method inspired by the biological brain \cite{bengio2017deep,franccois2017deep,lecun2015deep}. As stated in the literature, deep learning can be seen as an extension of shallow Neural Network models which have been around for many decades \cite{schmidhuber2015deep}, albeit the term deep learning with the current resurgence started in 2006 \cite{bengio2017deep,socher2012deep}. 

Techniques that incorporate deep learning have been steadily gaining in popularity \cite{bengio2017deep}. In this line of research, deep learning  have won numerous contests, in pattern and image recognition, and achieved promising results on different NLP tasks \cite{franccois2017deep,schmidhuber2015deep}. With deep learning algorithms, machines could learn good representations of data to help NLP tasks enormously. Specifically, deep learning seems to help in building constitutionality into Machine Learning models, just like human languages do to give meaning to complex ideas \cite{socher2012deep}. We can say that humans develop representations to enable learning and reasoning to achieve multiple tasks at hand like tackling the WSC, which indirectly relates with the schema hardness. In this regard, here, we train WinoReg within a Deep-Learning approach that is able to estimate the perceived human hardness indexes of Winograd schemas (see Figure \ref{fig:winoRegDL}).

\subsection{Data Enhancement via Crowdsourcing}
Although it is debatable \cite{Gary_Marcus}, it is widely accepted that deep learning killed feature engineering, which is time-consuming and brittle \cite{socher2012deep}. As stated in the literature, most of the time conventional ML techniques require considerable domain expertise for feature engineering \cite{lecun2015deep}.  On the other hand with deep learning, the amount of skill required for feature engineering reduces as the amount of training data increases \cite{bengio2017deep}. 

According to the literature, most approaches today that incorporate deep learning, succeed because we can provide them with the necessary resources, as it is widely accepted that to generalize better you have to do training on more data \cite{bengio2017deep}. Additionally, if we can provide deep learning with sufficient amount of data \cite{lecun2015deep}, it will also reduce the generalization error / over-fitting \cite{bengio2017deep}. 

In our case, the availability of training data is limited since we only have access on 143 schemas. In this regard, to increase Bender's training data we run an experiment on the \textbf{MicroWorkers (MW)} platform\footnote{\url{https://www.microworkers.com}}, which offers a reliable solution for various fields and research purposes \cite{peer2015beyond}; Amazon Turk, which was used in Bender's experiments, was not available in our region. Below, we will explain how we designed and ran our experiment along with our results.

\subsubsection{Dataset:}
According to the literature, multiple well-known datasets exist \cite{10.1007/978-3-030-35288-2_24,levesque2012winograd,morgenstern2016planning,kn:Rahman:2012:RCC:2390948.2391032,sakaguchi2020winogrande,wang2019glue}. 
Given that the original WSC286 dataset was used in Benders experiment, we designed our experiment based on the Rahman \& Ng's dataset (DPR). The main difference between the two datasets is the absence of questions in Rahman  \& Ng's schemas ---it only contains the definite pronoun. To match Bender's experiment we manually developed and added the necessary questions in all of the schemas. For the sake of simplicity, in our questionnaire, we use only the first half of each schema.

\subsubsection{Materials:}
For the design of the questionnaire we used LimeSurvey software from our lab server \footnote{\url{http://limesurvey.org}}.
All materials used in the experiment, including the schema halves used, are available online\footnote{\url{https://github.com/NicosCg/wsc-experiment}}.

\subsubsection{Participants:}
The questionnaire started in April 2020 and ran for two months. According to our results, a total of 306 participants from English speaking countries attempted and finished the task. Out of 429 participants who initially attempted the task, 115 did not finish ---the participants selected at least one answer but left before they completed the task. Furthermore, eight participants did not pass the testing phase (see \ref{design}). The total cost of our campaign was \$322. In the end, every schema half was answered by at least 30 participants.

\begin{figure}
	\centerline{\includegraphics[width=\columnwidth]{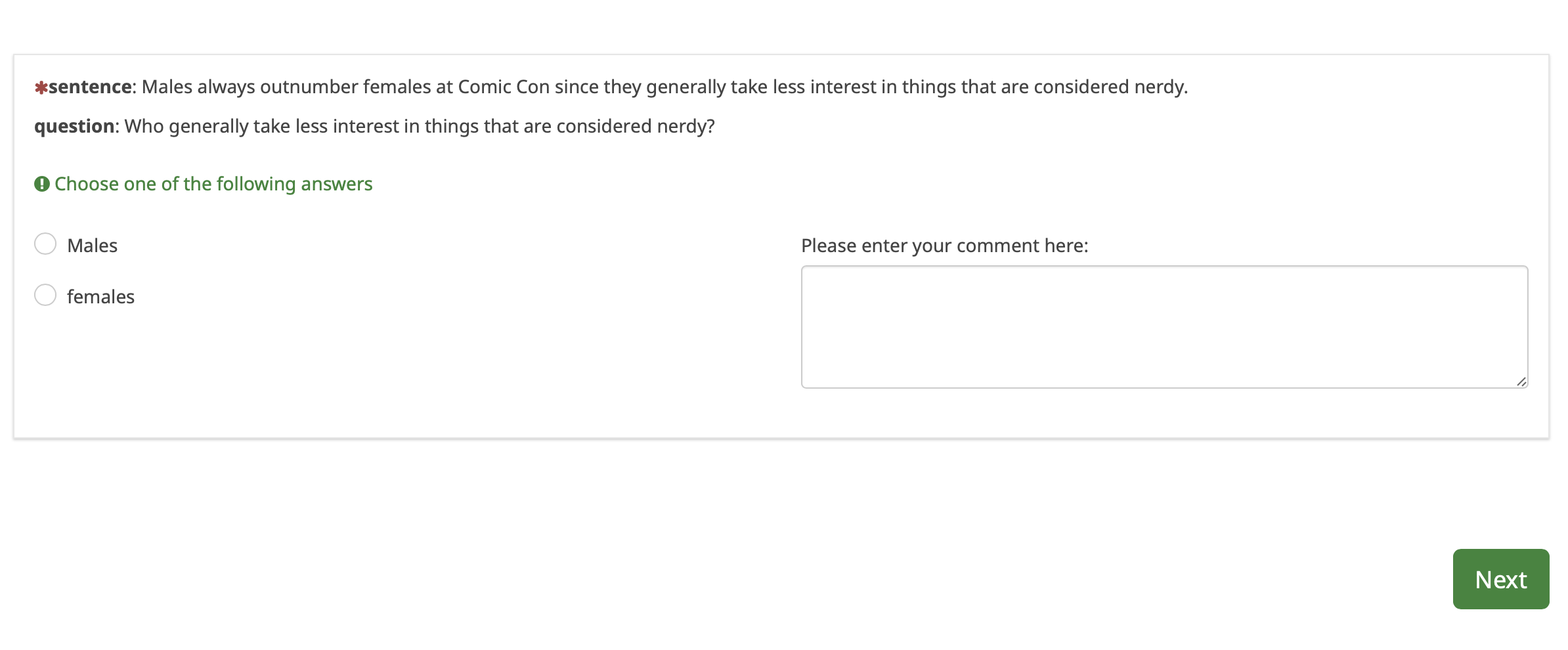}} 
	\caption{Screenshot of experiment window.}
	\label{fig:survey}
\end{figure}

\subsubsection{Design:} \label{design}
We built the questionnaire and posted the link on the Microworkers platform. A total of 943 schema halves were included, where, each half was displayed on a single screen. Each schema-half's sentence was displayed at the top, followed by the question, and the two possible answers that were displayed alongside (see Figure \ref{fig:survey}). Additionally, their was a comment section for participants to offer any comments they might have. All of the participants were informed that once the survey started, they could not change a submitted answer. Compared to Bender, our workers were not given an immediate feedback (correct or incorrect) after each trial, nor, by extension access on their updated score. 

Our questionnaire consisted of 10 sections that ran independently. Each section included 100 unique schema-halves except for the tenth, which included the last 43 schemas of the dataset. Each participant was allocated only one position, meaning that they were allowed to participate in only one section. 

Before taking the survey, each participant had to read a consent form to agree to participate. Next, they had to select their age, their English language literacy level, and pass a training phase to get familiarized with the task; in the training phase immediate feedback (correct/incorrect) was given to the participants. Ostensibly, instructions were given as a warning not to sacrifice accuracy for speed.

To avoid problems related to cheating we also included a number of test questions that were  randomly displayed among the other schemas. As dealing with cheating in crowdsourcing platforms is a major challenge \cite{10.1007/978-3-030-35288-2_24}, test questions were used to verify if a given worker indeed holds a particular skill \cite{christoforaki2014step,hirth2011anatomy}. Via an adaptive interjection of test questions at any time in any given place we aimed on the selection of the answers of really motivated participants. In this regard, in the end we selected only the answers of participants who scored at least 70\% on the test questions. Note that all participants were a priori informed about the test question mechanism. 

The testing phase consisted of 10 schema halves that were designed specifically to select, in the end, the answers of the best participants. According to Bender, a lot of schemas suffer from ambiguity, meaning that it is difficult even from humans to answer them \cite{kn:bender2015establishing}; this is related to the fact that the design of schemas is too difficult and troublesome \cite{morgenstern2016planning}. In this regard, the testing questions were designed in a way to directly show the correct pronoun antecedent (correct answer), without ambiguities. For instance, 
\textit{Sentence: Jane sings better than Susan because she is a professional. Question: Who is a professional? Answers: Jane, Susan. Correct Answer: Jane.}

\begin{figure}
	\centerline{\includegraphics[width=\columnwidth]{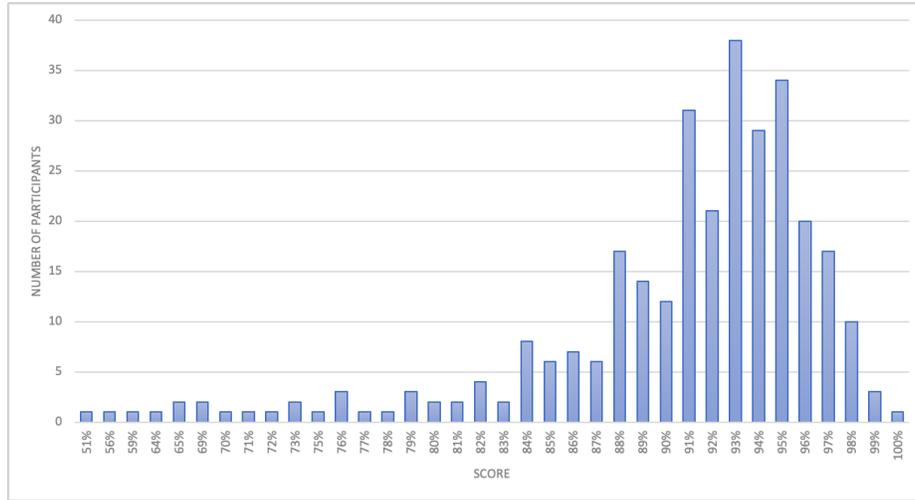}} 
	\caption{Questionnaire results: Distribution of scores.}
	\label{fig:participantsScore}
\end{figure}

\subsubsection{Results:} 
Based on the results, participants scored a mean accuracy of 91\% ($\sigma = 0.14$), taking an average of 17.9 ($\sigma = 1.09$) seconds to answer every schema-half (see Figure \ref{fig:participantsScore}). Our experimental results are in line with Bender's results \cite{kn:bender2015establishing}, meaning that the human adults can tackle the WSC with a mean of 91-92\%. The evidence we found supports Bender's results, meaning that this could serve as a baseline for human adult performance on the WSC. 
Furthermore, our results show that the two datasets do not differ significantly. Specifically, in Bender's work it was noted that the majority of Rahman and Ng's seems to be easy Winograd schemas. On the other hand, our results do not seem to confirm their observation. In fact, it seems that the hardness indexes of the two datasets are similar, meaning that human adults make the same effort to solve them, although the majority of the resend work in the literature believed that the DPR dataset is easier than the original dataset \cite{kocijan2020review}; This is in line with a recent work were it was shown that humans almost need the same effort to tackle schemas from the two challenges (96.5\% for the original WSC dataset and 95.2\% for the DPR dataset) \cite{sakaguchi2020winogrande} \footnote{See Table 6 on the relevant paper.}. On the other hand, this does not seem to be the case the machines, as it seems that they can tackle the DPR dataset with bigger success with a mean of 85\% compared to the WSC267 dataset where they score a mean of 71\% \cite{sakaguchi2020winogrande}; maybe this is due the fact that the DPR challenge instead of questions includes only the definite pronouns which makes the resolution to the machines easier to resolve, whilst on the other hand, the WSC dataset is more closely to a question answering 
challenge that might require bigger effort for machines to tackle it. 

\subsection{Fine-tuned dataset}
To get more data for our Deep-Learning approach we took the 943 schema halves of the Rahman \& Ng's dataset and added them to Bender's dataset along with their hardness indexes. 
To avoid having unbalanced data between the two datasets ---943 schema halves of the Rahman \& Ng's dataset over 286 schema halves of the original WSC dataset---, through oversampling, we increased the number of observations of the original dataset; these were copies of the existing schema halves, excluding the 100 schemas used for testing purposes. The whole process resulted in 1872 schema halves, which were used for training and testing purposes.

\subsection{WinoReg's Deep-Learning Architecture}
WinoReg's deep-learning architecture is based on LSTM networks (see Figure \ref{fig:winoRegDL}), an updated version of RNNs that are capable of learning long-term dependencies \cite{hochreiter1997long}; Specifically, LSTM networks may be also interpreted as something similar to a computer memory \cite{sundermeyer2012lstm}.
As stated in the literature, LSTM neural networks perform really well in the field of language modeling (LM) \cite{sundermeyer2012lstm} which can be used to solve various NLP tasks \cite{kocijan2019surprisingly}. A language model is an essential model that captures how meaningful sentences can be constructed from individual words, which, in our case, seems to relate to the hardness of schemas. In the absence of features, with LSTM networks WinoReg can learn the joint probability function of sequences of words in a given sentence \cite{bengio2003neural}, and at the same time, take into account all of the predecessor words  \cite{sundermeyer2015feedforward,sundermeyer2012lstm} to output the perceived human hardness index of any given schema.

Within this approach, WinoReg splits each examined schema-half to select the sentence, as this is the only input-value that is needed for our LSTM-based approach (see Figure \ref{fig:winoRegDL}). Next, it parses the examined sentence via spaCy dependency parser to remove the stop-words, since they often occur in abundance. Then, for every word in the sentence it returns its lemmatization as a way to determine possible relations between common-words. The final step is to feed the parsed sentence into the model to retrieve its hardness index.

\section{Experimental Evaluation}
In this section, we present our results by applying the methodology described in this paper. In this regard, we undertook several experiments to investigate if WinoReg can be used to automatically differentiate between Winograd schemas based on their perceived hardness for humans. We start by presenting WinoReg results based on the Random-Forest approach and continue with the LSTM-based approach. Both experiments ran on a laptop-computer (MacBook Pro 2018) with 2.2 GHz 6-Core Intel Core i7 CPU, 16GB's RAM, Radeon Pro 555X GPU with 4GB of GDDR5.

\subsection{Random-Forest Approach}
Here, by using the data from Bender's study \cite{kn:bender2015establishing}, we examine whether the performance of the Random-Forest approach can be predictive of the hardness of the WSC instances for humans. The results are reported on the testing set, which comprises 30\% of the original WSC dataset (286 schema halves), expressed in terms of accuracy and correlation coefficient. For comparison purposes, the testing set is identical to the one that was used in our first work \cite{GCAI-2018:Data_Driven_Metric_of_Hardness}. According to Bender's results, the human adult bar on the testing set (100 schemas halves) is 91\%.

\begin{table}[h]
	\centering

	\begin{tabular}{|c|c|c|c|c|}
		\hline
		System & Correlation Coefficient & Accuracy \tabularnewline
		\hline
		\hline
		Fixed Baseline & -1 & 90.87 \tabularnewline
		\hline
		Wikisense-based & 0.22 & 77 \tabularnewline
		\hline
		WinoReg-RF & 0.33 & 91.64 \tabularnewline
		\hline
	\end{tabular}
	
	\caption{Results of the Fixed Baseline, the Wikisense-based hardness, and WinoReg, which was trained based on the Random-Forest approach.}
	\label{Table: hardnessresutlts}
\end{table}

\begin{figure}
	\centerline{\includegraphics[width=\columnwidth]{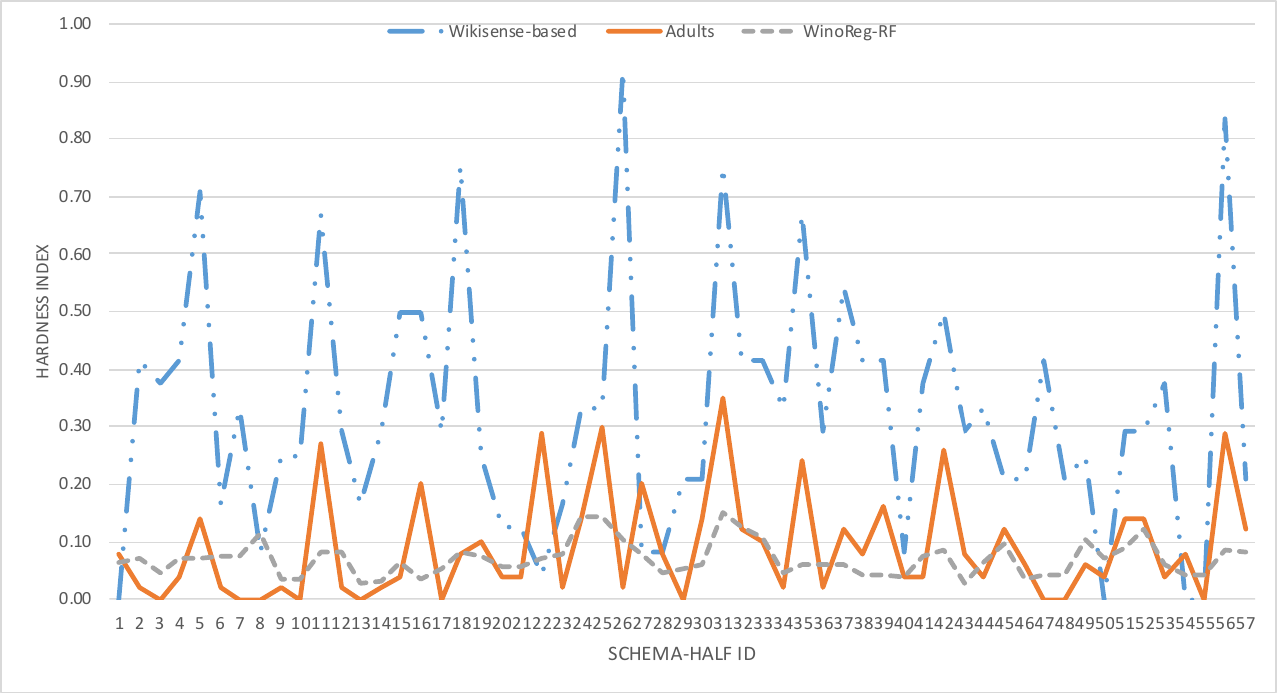}} 
	\caption{Variability of WinoReg and Wikisense-based hardness-index across the 57 WSC instances on which the Wikisense-based approach originally was computed (in relation to the variability of the human hardness-index).}
	\label{fig:randomF57all}
\end{figure}

\begin{figure}
	\centerline{\includegraphics[width=\columnwidth]{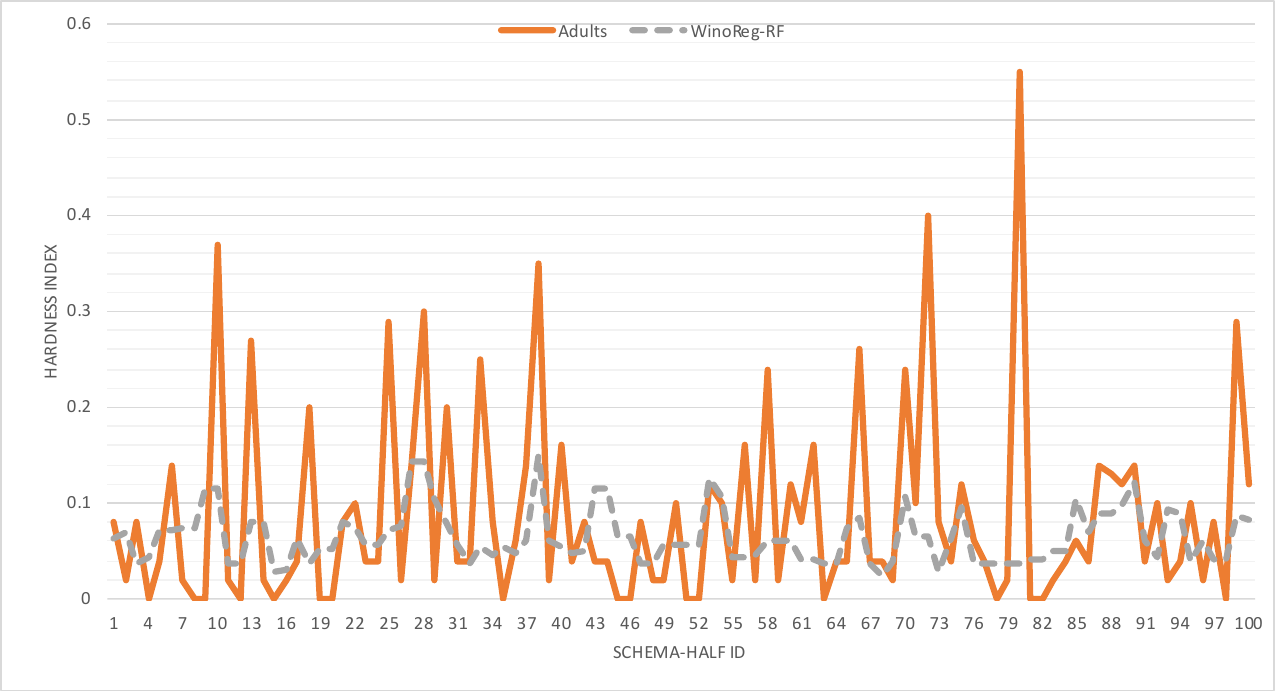}} 
	\caption{Variability of the WinoReg hardness-index and the perceived human hardness-index across our testing set (100 schema halves). WinoReg is trained based on the Random-Forest approach.}
	\label{fig:RFresults}
\end{figure}

\subsubsection{Results and Discussion} \hfill\\

\vskip 0.1in \noindent
\textbf{The fixed baseline:} For comparison purposes, we trained our Random Forest algorithm with only one feature, which is the human adult bar of 91\% and, like in our first work we tested it on the first 100 schema halves. Not surprisingly, our results show an achievement of 90.87\%, but with a Correlation Coefficient with the adults results, of -1 (see Table \ref{Table: hardnessresutlts}).

\vskip 0.1in \noindent
\textbf{Wikisense-based Hardness:} Recall that the Wikisense-based hardness is able to return results only for 57\% of the examined schemas with a correlation coefficient of 38\% \cite{GCAI-2018:Data_Driven_Metric_of_Hardness}. It seems that Wikisense, the engine behind the system, was unable to output the necessary keywords to search the Wikipedia corpus. Given that the human adult bar on our testing set is 91\% (for the unresolved schemas), we can assume that Wikisense-based achieves an accuracy of 77\% on all of the remain schemas, with a correlation coefficient of 22\% (see Fig. \ref{fig:randomF57all}).

\vskip 0.1in \noindent
\textbf{WinoReg:} The general picture emerging from the analysis is that WinoReg can achieve an accuracy of 91.64\%, significantly outperforming the Wikisense-based approach by 14.64\% in accuracy and by 11\% in correlation coefficient (see Fig. \ref{fig:RFresults}). To make a better comparison between WinoReg and Wikisense-based approach, we compared the two methods only on the 57 schema halves the Wikisense-based system was able to resolve. In this regard, the correlation coefficient of WinoReg and humans rises to 47\%, which is 9 percentage points bigger than what the Wikisense-based system was able to achieve (38\%). 

Taken altogether, the data presented here provide evidence that the performance of WinoReg, which is based on the random forest algorithm, \emph{varies} across WSC instances in a manner that resembles the variability of the human performance more closely than what previous systems could achieve. This can be seen in both Fig. \ref{fig:randomF57all} and Fig. \ref{fig:RFresults} that depict how the computed hardness index and the human hardness index vary across WSC instances, suggesting that indeed, certain WSC instances that are easier or harder for humans are accordingly labeled as such by WinoReg.

\subsubsection{Speed Analysis}\hfill\\
Given that the hardness index plays an important role in the quality of the developed schemas and that there are crowdsourcing and machine driven approaches that already leverage the Wikisense-based hardness mechanism \cite{10.1007/978-3-030-35288-2_24}, it is crucial to have access to the hardness index without delays. In this regard, we performed a speed analysis to show how fast WinoReg can provide us with the hardness index of Winograd schema-halves. Compared to the Wikisense-based approach, which requires on average 8 hours for every schema half, it was found that WinoReg can return the hardness index of a schema half, on average, in 1.6 minutes; this is the time needed for the estimation of the required features that are fed to the Random-Forest model. The results ultimately show that WinoReg can deliver the hardness index of schemas 300 times faster than the Wikisense-based approach.

\subsubsection{Feature Analysis} \hfill\\

\begin{figure}
	\centerline{\includegraphics[width=\columnwidth]{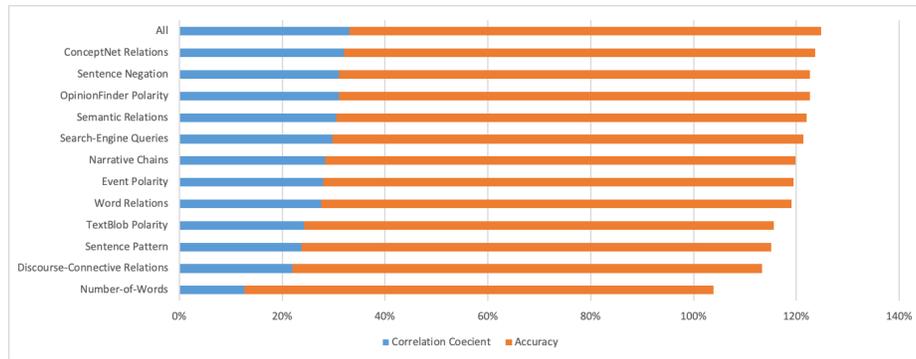}} 
	\caption{Results of feature decrement experiments. We can see the performance of the model trained on all types of features except for the one shown in that row.}
	\label{fig:featureimpo}
\end{figure}

Here, we present the results that were obtained from an analysis of the features used to train our Random Forest model. Consequently, as shown in Fig. \ref{fig:featureimpo}, where each element on the Y axis presents the performance of WinoReg trained on all types of features except for the one shown, the correlation coefficient drops significantly whichever feature is removed. In this regard, the results provide evidence of the importance of all feature types. 

The results show that the Number-of-Words, the Discourse-Connective-Relations, the Sentence-Pattern, the TextBlob-Polarity, and the Word-Relations are the most important features. This is in line with previous studies, where it was shown that features like the sentence length, sentence pattern and word relations play an important role on both the quality of the schemas and the tackle of the challenge \cite{10.1007/978-3-030-35288-2_24,icaart20,kn:Rahman:2012:RCC:2390948.2391032}.  

Regarding the TextBlob-Polarity, our results show that it is better in capturing the polarity context than the other polarity features, which was unexpected as it is not commonly used in the literature. Regarding the OpinionFinder previous works have stated, that this might happen because it was trained on a completely different training set \cite{kn:Rahman:2012:RCC:2390948.2391032}. 

Contrary to our expectations, and unlike what other studies have mentioned \cite{kn:Rahman:2012:RCC:2390948.2391032}, 
Search-Engine-based features are not among the most useful features. We believe that this might have happened because of changes in the Google search algorithm, which might have led to different results. Additionally, contrary to other works \cite{kn:budukh2013intelligent}, it seems that ConceptNet-Relations is not among the most useful features. Maybe its similarity factor cannot easily capture the semantics of each sentence. Lastly, it seems that the Negation-Feature is among the features that offer the least, which might be attributed to the fact that our dependency parser was able to determine if negation exists in only 41\% of the schema halves.

\subsection{LSTM-based Approach}
In this section, we present our results by applying the LSTM-based approach described in the previous sections. Within our experiments, we examine whether this a priori appropriateness of the LSTM-based approach can be predictive of the hardness of the WSC instances for humans. The results are expressed in terms of accuracy and correlation coefficient.
For comparison purposes, the testing set is identical to the one used in both the Random Forest and the Wikisense-based approach.

The optimal values for hyper-parameters used, which are vital to enhancing the training result of a neural network model \cite{inproceedingsLe}, were determined through trial-and-error. To build and train our model, we used the Keras functional API \cite{francois2017deep}. Our model was compiled with the ``Adamax" optimizer and the ``mean\_absolute\_error" loss function to compute the mean of the absolute difference between labels and predictions. In this regard, the regression loss function represents the measure of success for the task at hand to predict the hardness-index of any schema half.

We started with the Sequential model API, where model layers were created and added to it. Initially, we added an embedding layer to associate vectors with words. In this regard, we considered all the words in our dataset (input dim= 3648), with a maximum of each half’s sentence of 50 words (output dim=50). Next, we added our LSTM layer, consisting of eighty-seven units (neurons), and, to prevent overfitting, we used a dropout layer (0.2) to ignore randomly selected neurons during the training process. Additionally, we used a recurrent dropout of 0.2 to mask the connections between the recurrent units. Finally, we added a single unit layer to reduce our LSTM network's shape to match our desire output (hardness score prediction).

We have also examined transfer-learning to train our model with better generalization properties. As stated in the literature, several transfer-learning methods significantly improved a wide range of NLP tasks \cite{ruder2019transfer}. To improve the accuracy of our model, we have tested two well-known datasets, Glove (glove.6B.50d to glove.6B.300d) \cite{pennington2014glove}, and fastText (cc.en.300.vec) \cite{joulin2016fasttext}, albeit without any success. As stated in the literature, this is something that might happen in natural language analysis tasks, maybe because these kinds of pre-rained embeddings suffer from a paucity of data \cite{qi2018and}.
	
Finally, for our training and testing purposes, we split the dataset into a training and a validation set (validation\_split=0.3; train on 1310 samples, validate on 562 samples) and tested the resulting model on the testing dataset (100 samples).

\subsubsection{Results and Discussion} \hfill\\

\begin{table}[h]
	\centering

	\begin{tabular}{|c|c|c|c|c|}
		\hline
		System & MAE & Correlation Coefficient & Accuracy \tabularnewline
		\hline
		\hline
		Wikisense-based & 23 & 0.22 & 77 \tabularnewline
		\hline
	    WinoReg-RF & 8.36 & 0.33 & 91.64 \tabularnewline
		\hline
	    WinoReg-DL (LSTM-based) & 0.673 & 0.39 & 93.27 \tabularnewline
		\hline
	\end{tabular}
	
	\caption{Results of the Fixed Baseline, the Wikisense-based hardness, and WinoReg based on both the Random-Forest and the LSTM-based Approach.}
	\label{Table: results}
\end{table}

\begin{figure}
	\centerline{\includegraphics[width=\columnwidth]{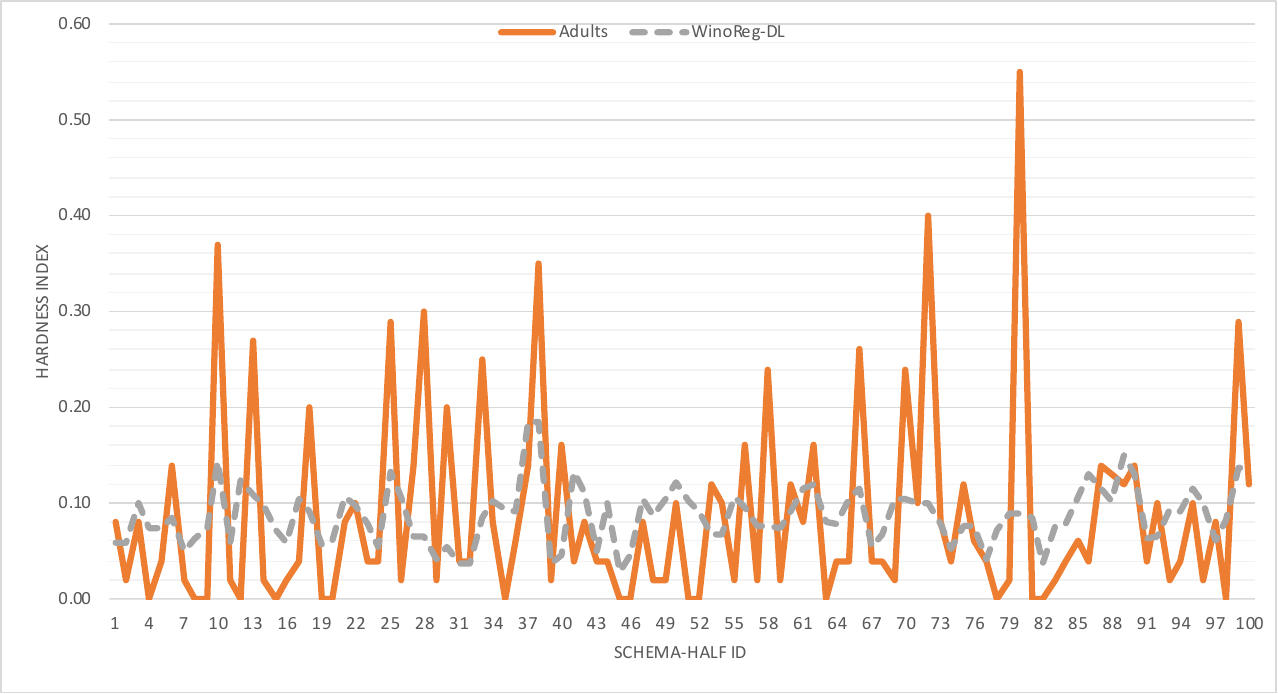}} 
	\caption{Variability of the WinoReg hardness-index and the perceived human hardness-index across our testing set (100 schema halves). WinoReg is trained based on the LSTM-based approach.}
	\label{fig:DLresults}
\end{figure}

\begin{figure}
	\centerline{\includegraphics[width=\columnwidth]{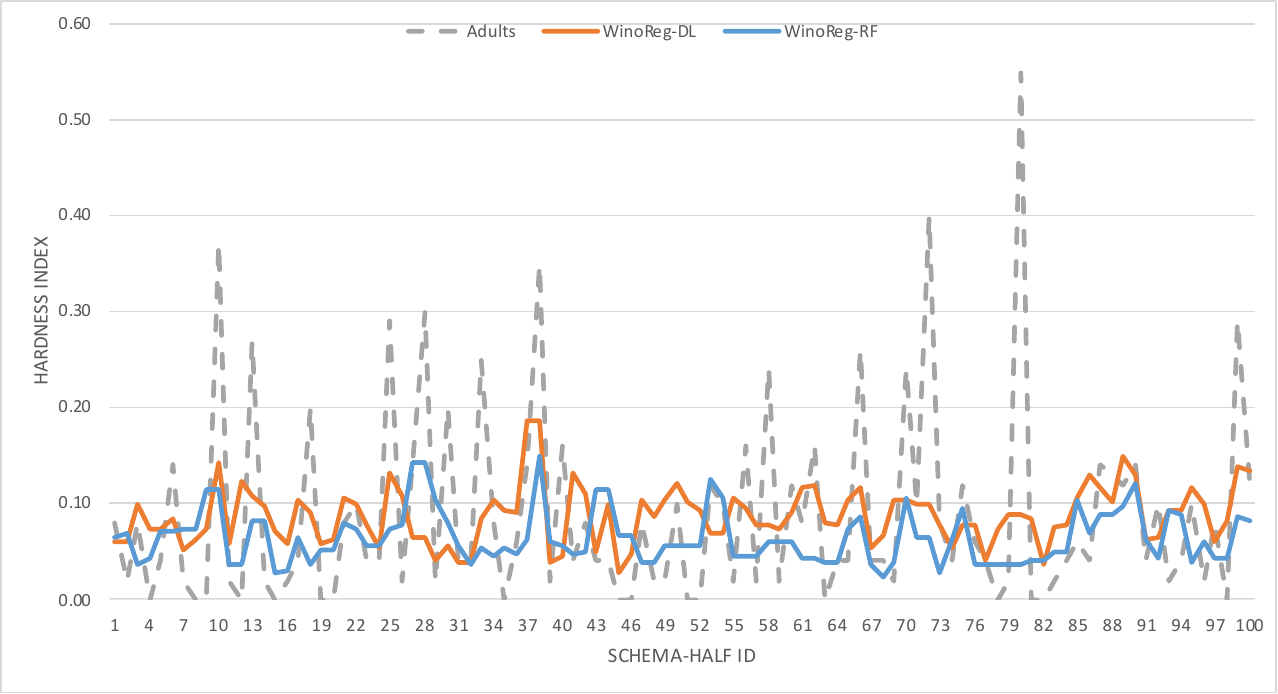}} 
	\caption{Variability of WinoReg approaches, in relation to the perceived human hardness-index across our testing set (100 schema halves).}
	\label{fig:winoregRFDL}
\end{figure}

Our tests show that there is a positive correlation between WinoReg results and the perceived human hardness-indexes, across the Winograd schemas (see Table \ref{Table: results}).
Specifically, within the LSTM-based approach, WinoReg can achieve an accuracy of 93.27\% with a correlation coefficient of 39\%.

Compared to Wikisense-based approach, WinoReg, within the LSTM-based approach can achieve a higher correlation coefficient of 17\%. Additionally, if we compare the two systems on the 57\% of the schemas the Wikisense-based approach was able to solve, the correlation-coefficient difference rises to 10\%, in favor of WinoReg (38\% vs 48\%).

As shown in Table \ref{Table: results}, the LSTM-based approach has an advantage over the Random-Forest approach. Our results highlighted that WinoReg results correlate better to human adult results in the case of LSTM-based than the Random-Forest approach. Specifically, the former outperforms the latter by 2\% in accuracy and 6\% in correlation coefficient. Additionally, the fact that the LSTM-based approach does not require feature engineering, offers it a compelling advantage over the Random-Forest approach.

The general picture emerging from the analysis is that the performance of WinoReg, when trained based on the LSTM-based approach, \emph{varies} across Winograd schemas in a way that resembles the variability of the human performance more closely than what other approaches could achieve (see Fig. \ref{fig:DLresults}). In this regard, certain WSC instances that are easier or harder for humans are respectively identified as such by WinoReg. Broadly speaking, both approaches have a compelling advantage over the Wikisense-based approach. Specifically, further analysis we undertook showed that both WinoReg approaches correlate positively with a correlation coefficient of 22\%, suggesting that schemas that are easier or harder for the Random-Forest approach are accordingly labeled as such by the LSTM-based approach (see Figure \ref{fig:winoregRFDL}).

\vskip 0.1in \noindent
\textbf{Speed Analysis:}
Recall that having access to schema hardness indexes is crucial for both the quality of the schemas and the CAPTCHA service. In this regard, we performed a speed analysis to identify how fast the Deep-Learning approach can provide us with results. Our analysis revealed that within the LSTM-based approach WinoReg can return results, on average, in 1.6 msec for any given schema, which means 60 thousand times faster the Random Forest and 18 million times faster the Wikisense-based approach. This means that WinoReg, within the LSTM-based approach, is able to return results in real-time with no further delays.

\section{Related Work}
The work presented here is not unrelated to a recent work, where we have demonstrated the possibility of using crowdsourced workers for the development of Winograd schemas \cite{10.1007/978-3-030-35288-2_24}. WinoFlexi is an online collaboration platform where workers collaborate with the help of various tools that enhance the schema development process. One of this tools is the Wikisense-based approach, which indirectly helps workers develop schemas of various hardness indexes, albeit with big delays. For instance, if the majority of the developed schemas of a crowdworker are considered easy, WinoFlexi prompts them to develop schemas that are harder to solve. Recall that this might lead to high development costs because the WinoFlexi's notification mechanism depends on the Wikisense-based approach, which is time-consuming. In this regard, if we replace the Wikisense-based approach with WinoReg, it will further reduce the schema development costs. With WinoReg (DL), WinoFlexi will be able to access the hardness-indexes of schemas in real-time.  

In another work, which is relevant to the previous one \cite{icaart20}, we have designed a system that offers a full pipeline for automated or semi-automated design of schemas. At the same time,  it considerably helps humans in the schema development task. Evidence from that study has shown that the developed system is able to automatically design large amounts of schemas, albeit of lower quality to that developed by humans \cite{10.1007/978-3-030-35288-2_24}. On the other hand, it was shown that the system is able to considerably motivate and inspire humans for the development of high-quality schemas. In this regard, WinoReg could be used to help humans develop schemas of various hardness-indexes.

The first and only Winograd schema challenge was organized back in 2016, along with the IJCAI 2016 conference. According to the organizers, the design of schemas was found to be too troublesome and difficult to be handled at regular intervals for short periods of time \cite{morgenstern2016planning}. Through an experiment the organizers evaluated the hardness of each examined schema (consisted of 89 problems with 9 subjects) which helped in the organization of the challenge. Their results have shown a 91\% of achievement which is in line with ours and Bender's results. They categorized their schemas according to the number of the correct answers given by participants, which resembles the way WinoReg works. In this regard, instead of using human participants, they could use WinoReg mechanisms which could save them time and money ---Although participants were paid for their participation, the authors did not mention the amount payed.

In an earlier work we have demonstrated how Winograd schemas can form a novel form of CAPTCHAs \cite{GCAI-2018:Using_Winograd_Schema_Challenge}. Specifically, by providing motivation for a detailed form of WSC-based CAPTCHAs we have shown that this kind of CAPTCHAs are equally entertaining and useful like the other form of CAPTCHAs. The WSC-based CAPTCHAs are generated as the means to identify humans from bots, and at the same time to prevent automated processes from performing illicit actions. WinoReg can contribute by organizing schemas according to their perceived hardness for humans to be displayed accordingly by the CAPTCHA service.

A plethora of works in the literature focused on the development of systems for tackling the WSC. In this regard, Emami et al. \cite{emami2018knowledge} developed a rule-based system that, by focusing on knowledge-hunting on the Web, was able to achieve better than 57\% accuracy on the original WSC problem (WSC273). According to Kocijan et al. \cite{kocijan2020review}, this was the first approach to achieve better than chance accuracy. Based on well-formed search engine queries, they search the WWW to extract knowledge that helps them to identify the correct pronoun target of each schema half. 

In a follow-up work, Emami et al. \cite{emami2019knowref} introduced the KNOWREF Coreference Task, which provides 8,724 difficult pronoun disambiguation problems extracted from various sources (English Wikipedia, OpenSubtitles, and Reddit). According to the authors, their dataset is more challenging to resolve than other similar datasets.  This is done by using various techniques to construct a human-labeled corpus of Winograd-like text that targets systems' ability to reason about a situation described in the context. The authors show that training coreference models on KNOWREF improve their ability to do better representations of the context, linked to reducing gender bias. In support of their claim, experiments showed that state-or-the-art models rely on the gender or number of candidate antecedents to make a decision.

Kocijan et al. \cite{kocijan2019wikicrem} introduced a new dataset called WIKICREM, developed by masking repeated occurrences of personal names. According to the authors, language models, which are very promising for tackling the pronoun resolution challenge, when pre-trained on an extensive collection of raw-text, can be easily fine-tuned on a specific task using much less training data. By introducing WIKICREM, which consists of 2.4 million quickly extended examples, Kocijan et al. address the lack of large training sets, which could be used with language models. Basically, by finding passages of text where a personal name appears at least twice, they mask one of its non-first occurrences. Experiments show that fine-tuning the BERT \cite{devlin2019bert} language model with WIKICREM consistently improves the model (84.8 \% accuracy on DPR and 71.8 on WSC273).

Trinh et al., \cite{trinh2018simple}, via an ensemble of LSTM language models, pre-trained on a large corpus of text (LM-1-Billion, CommonCrawl6, SQuAD, and Gutenberg Books) tackled the original WSC (WSC273) with an accuracy of 63.7\%. Their approach, which was one of the first to use a pre-trained language model \cite{kocijan2020review}, selected the best pronoun target, which was the target that formed the best English sentence. In this regard, in each schema half's sentence, they substitute the pronoun with the candidates and then use Language Model to estimate the two sentences' probability.

Sakaguchi et al., \cite{sakaguchi2020winogrande} tested if recent advances in neural Language Models, which reportedly reached around 90\% accuracy in WSC-like instances, have acquired commonsense capabilities or they rely on spurious biases found in the training datasets. In this regard, they developed a sizeable WSC-like dataset (WinoGrande), which consists of 44 thousand examples, collected via crowdsourcing on Amazon Mechanical Turk. To help workers develop non-identical schemas, they primed them with a randomly chosen topic from a WikiHow article \cite{kocijan2020review}. They also filtered-out a smaller de-biased dataset (called WinoGrande-debiased), which consists of 12,282 instances. In this regard, based on a fine-tuned RoBERTa (Robustly Optimized BERT-Pretraining Approach) language model \cite{liu2019roberta}, they gained contextualized embeddings for each instance. Next, they used those embeddings to discard the top instances that were correctly resolved by more than 75\% of the classifiers trained on the embeddings. According to their results, when training on the WinoGrande dataset, RoBERTa, which is trained initially on a vast dataset of over 160GB of uncompressed text, achieves new state-of-the-art results on the original WSC (90.1\%). WinoGrande has its lowest-achieved rate of 85\% on the KNOWREF dataset. Although this shows the strength of the WinoGrande dataset when used as a resource for transfer-learning, it also raises the concern that they are likely to be overestimating the true capabilities of machine commonsense \cite{sakaguchi2020winogrande}. According to Lin et al. \cite{lin2020tttttackling} despite careful controls, the WinoGrande challenge might contain incidental biases that these more sophisticated models can exploit.

Regarding biases in corpora, Webster et al. \cite{webster2018mind} argued that existing corpora do not capture ambiguous pronouns meaning that they are mostly based on gender bias to tackle coreference resolution problems. For instance, in experiments, they found out that gender bias in existing corpora favors masculine entities.  According to the authors, Winograd schemas are closely related to their work as they contain ambiguous pronouns. To address issues related to gender bias, they developed a corpus (GAP) that consists of about 8,908 ambiguous pronoun–name pairs derived from Wikipedia (development and testing set of 4,000 examples, and 908 examples for parameter tuning). Coreference resolvers trained and tested on GAP were found to struggle, showing that ambiguous pronoun resolution remains a challenge.

According to Brown et al. \cite{brown2020language} few-shot learning is an advantageous technique in machine learning when there is a very small amount of data available. In this regard, in recent research, which incorporates GPT-3, a vast language model with over 175 billion parameters, they test if it is possible to tackle various NLP challenges like the WSC. Specifically, they examine three kinds of training: i) zero-shot, which means that GPT-3 is given only the description of the challenge with zero examples of the task; ii) one-shot, which means that along with the description of the task at hand, the model is given only one example of the challenge; iii) few-shot, which means that the model sees few examples of the challenge.  Regarding the WSC,  GPT-3 is tested on the original set (WSC 273), where it achieves 88.3\%, 89.7\%, and 88.6\% in the zero-shot, one-shot, and few-shot settings; On the WinoGrande dataset, the model achieves 70.2\%, 73.2 \%, and 77.7\% in the zero-shot, one-shot, and few-shot settings. On the other hand, further analysis they performed showed that 45\% of the Winograd schemas were presented in the data used for the training of GPT-3 (Common Crawl + several curated high-quality datasets), which led to a 2.6\% decrease in performance on the clean subset. 

Lin et al.\cite{lin2020tttttackling}, via a T5 encoder-decoder model, which is trained based on the Common-Crawl-based data, tackled the WinoGrande dataset with an accuracy of 77\%. According to the authors, encoder-decoder models can tackle, among others, comprehension and text generation tasks. For fine-tuning purposes, each example is split in half to end-up with a two-statement problem (called the source). Each statement contains the hypothesis and the premise, referring to either the first or the second candidate. Finally, the correct statement is labeled with the entailment label while the other with contradiction.  At inference (test) time, each schema is decomposed the same as in the fine-tuning phase.  

\section{Conclusion and Future Work}
This paper has investigated the possibility of building a system that can output the perceived human hardness-index of any Winograd schema in the shortest time possible. Our results have shown that this is possible via the training of a system that is based on two different approaches, namely, the Random-Forest and the Deep-Learning (LSTM-based) approach. We have provided evidence of that by comparing WinoReg results with two studies, one from the literature \cite{kn:bender2015establishing} and one that we designed and undertook. Results have shown that WinoReg results correlate positively with human results. In particular, results have shown that with the Random-Forest approach we can achieve 91.64\% of accuracy with 33\% correlation coefficient, whereas with the Deep-Learning approach 93.27\% of accuracy with 39\% correlation coefficient. Even though the results of the two approaches seem close, the strong benefit of the Deep-Learning approach lies in the response time of the model, which is 60 thousand times faster than the Random-Forest model.

WinoReg can be used by researchers or challenge organizers to group schemas in terms of their perceived human hardness indexes. Specifically, WinoReg can be used by CAPTCHA organizers to ensure that the generated schemas are not overly demanding for human users. Additionally, WinoReg can be used in systems that pursue the development of Winograd schemas from scratch, like in \cite{10.1007/978-3-030-35288-2_24,icaart20}, to ensure that a variety of schemas would be developed. We suggest that future studies should examine the impact of systems like WinoReg in other AI fields. For instance, in the field of machine translation, systems like WinoReg could be used to identify sentences that are harder to translate, in order to acquire better feedback from people. In this regard, WinoReg can help with the problem many translation services face, of where to focus their attention to make end-users aware of the quality \cite{specia2009estimating}.

\section*{Acknowledgments}
This work was supported by funding from the EU's Horizon 2020 Research and Innovation Programme under grant agreements no. 739578 and no. 823783, and from the Government of the Republic of Cyprus through the Directorate General for European Programmes, Coordination, and Development.

%
%
%
%
\bibliographystyle{splncs04}
\bibliography{isaakZjuj}

\begin{thebibliography}{10}
\providecommand{\url}[1]{\texttt{#1}}
\providecommand{\urlprefix}{URL }
\providecommand{\doi}[1]{https://doi.org/#1}

\bibitem{baker1998berkeley}
Baker, C.F., Fillmore, C.J., Lowe, J.B.: {The Berkeley Framenet Project}. In:
  Proceedings of the 17th international conference on Computational
  linguistics-Volume 1. pp. 86--90. Association for Computational Linguistics
  (1998)

\bibitem{kn:bender2015establishing}
Bender, D.: {Establishing a Human Baseline for the Winograd Schema Challenge.}
  In: MAICS. pp. 39--45 (2015)

\bibitem{bengio2003neural}
Bengio, Y., Ducharme, R., Vincent, P., Jauvin, C.: {A Neural Probabilistic
  Language Model}. Journal of machine learning research  \textbf{3}(Feb),
  1137--1155 (2003)

\bibitem{bengio2017deep}
Bengio, Y., Goodfellow, I., Courville, A.: Deep learning, vol.~1. MIT press
  (2017)

\bibitem{bhagavatula2019abductive}
Bhagavatula, C., Bras, R.L., Malaviya, C., Sakaguchi, K., Holtzman, A.,
  Rashkin, H., Downey, D., Yih, S.W.t., Choi, Y.: {Abductive Commonsense
  Reasoning}. arXiv preprint arXiv:1908.05739  (2019)

\bibitem{blanco2011some}
Blanco, E., Moldovan, D.: {Some Issues on Detecting Negation From Text}. In:
  Twenty-Fourth International FLAIRS Conference (2011)

\bibitem{breiman2001random}
Breiman, L.: Random forests. Machine learning  \textbf{45}(1),  5--32 (2001)

\bibitem{brown2020language}
Brown, T.B., Mann, B., Ryder, N., Subbiah, M., Kaplan, J., Dhariwal, P.,
  Neelakantan, A., Shyam, P., Sastry, G., Askell, A., Agarwal, S.,
  Herbert-Voss, A., Krueger, G., Henighan, T., Child, R., Ramesh, A., Ziegler,
  D.M., Wu, J., Winter, C., Hesse, C., Chen, M., Sigler, E., Litwin, M., Gray,
  S., Chess, B., Clark, J., Berner, C., McCandlish, S., Radford, A., Sutskever,
  I., Amodei, D.: {Language Models are Few-Shot Learners} (2020)

\bibitem{kn:budukh2013intelligent}
Budukh, T.U.: {An Intelligent Co-reference Resolver for Winograd Schema
  Sentences Containing Resolved Semantic Entities.} Master's thesis, Arizona
  State University (2013)

\bibitem{kn:chambers2008unsupervised}
Chambers, N., Jurafsky, D.: {Unsupervised Learning of Narrative Event Chains.}
  In: ACL. vol. 94305, pp. 789--797 (2008)

\bibitem{christoforaki2014step}
Christoforaki, M., Ipeirotis, P.: {Step: A Scalable Testing and Evaluation
  Platform}. In: Proceedings of the 2nd AAAI Conference on Human Computation
  and Crowdsourcing (2014)

\bibitem{eniac}
Cozman, F., Munhoz, H.: {The Winograd Schemas from Hell}. In: Anais do XVII
  Encontro Nacional de Inteligência Artificial e Computacional. pp. 531--542.
  SBC, Porto Alegre, RS, Brasil (2020). \doi{10.5753/eniac.2020.12157},
  \url{https://sol.sbc.org.br/index.php/eniac/article/view/12157}

\bibitem{dagan2005pascal}
Dagan, I., Glickman, O., Magnini, B.: {The Pascal Recognising Textual
  Entailment Challenge}. In: Machine Learning Challenges Workshop. pp.
  177--190. Springer (2005)

\bibitem{devlin2019bert}
Devlin, J., Chang, M.W., Lee, K., Toutanova, K.: {BERT: Pre-training of Deep
  Bidirectional Transformers for Language Understanding} (2019)

\bibitem{emami2018knowledge}
Emami, A., De~La~Cruz, N., Trischler, A., Suleman, K., Cheung, J.C.K.: {A
  Knowledge Hunting Framework for Common Sense Reasoning}. arXiv preprint
  arXiv:1810.01375  (2018)

\bibitem{emami2019knowref}
Emami, A., Trichelair, P., Trischler, A., Suleman, K., Schulz, H., Cheung,
  J.C.K.: The knowref coreference corpus: Removing gender and number cues for
  difficult pronominal anaphora resolution. In: Proceedings of the 57th Annual
  Meeting of the Association for Computational Linguistics. pp. 3952--3961
  (2019)

\bibitem{franccois2017deep}
Fran{\c{c}}ois, C.: {Deep Learning with Python} (2017)

\bibitem{francois2017deep}
Francois, C.: {Deep Learning with Python} (2017)

\bibitem{fry2018hello}
Fry, H.: Hello World: How to be Human in the Age of the Machine. Random House
  (2018)

\bibitem{Gary_Marcus}
{Gary Marcus}: {Beyond Deep Learning with Gary Marcus}. [online] (2019),
  \url{https://hbr.org/podcast/2019/10/beyond-deep-learning-with-gary-marcus}

\bibitem{hassan2010identifying}
Hassan, A., Radev, D.: {Identifying Text Polarity Using Random Walks}. In:
  Proceedings of the 48th Annual Meeting of the Association for Computational
  Linguistics. pp. 395--403. Association for Computational Linguistics (2010)

\bibitem{hirth2011anatomy}
Hirth, M., Ho{\ss}feld, T., Tran-Gia, P.: {Anatomy of a Crowdsourcing Platform
  --- Using the Example of microworkers.com}. In: Proceedings of the 5th
  International Conference on Innovative Mobile and Internet Services in
  Ubiquitous Computing. pp. 322--329. IEEE (2011)

\bibitem{hochreiter1997long}
Hochreiter, S., Schmidhuber, J.: {Long Short-Term Memory}. Neural computation
  \textbf{9}(8),  1735--1780 (1997)

\bibitem{Isaak2016}
Isaak, N., Michael, L.: {Tackling the Winograd Schema Challenge Through Machine
  Logical Inferences.} In: Pearce, D., Pinto, H.S. (eds.) STAIRS. Frontiers in
  Artificial Intelligence and Applications, vol.~284, pp. 75--86. IOS Press
  (2016),
  \url{http://dblp.uni-trier.de/db/conf/stairs/stairs2016.html#IsaakM16}

\bibitem{GCAI-2018:Data_Driven_Metric_of_Hardness}
Isaak, N., Michael, L.: {A Data-Driven Metric of Hardness for WSC Sentences}.
  In: Lee, D., Steen, A., Walsh, T. (eds.) GCAI-2018. 4th Global Conference on
  Artificial Intelligence. EPiC Series in Computing, vol.~55, pp. 107--120.
  EasyChair (2018). \doi{10.29007/398z},
  \url{https://easychair.org/publications/paper/nRrp}

\bibitem{GCAI-2018:Using_Winograd_Schema_Challenge}
Isaak, N., Michael, L.: {Using the Winograd Schema Challenge as a CAPTCHA}. In:
  Lee, D., Steen, A., Walsh, T. (eds.) GCAI-2018. 4th Global Conference on
  Artificial Intelligence. EPiC Series in Computing, vol.~55, pp. 93--106.
  EasyChair (2018). \doi{10.29007/rnk8},
  \url{https://easychair.org/publications/paper/pV9V}

\bibitem{10.1007/978-3-030-35288-2_24}
Isaak, N., Michael, L.: {WinoFlexi: A Crowdsourcing Platform for the
  Development of Winograd Schemas}. In: Liu, J., Bailey, J. (eds.) AI 2019:
  Advances in Artificial Intelligence. pp. 289--302. Springer International
  Publishing, Cham (2019)

\bibitem{icaart20}
Isaak., N., Michael., L.: {Winventor: A Machine-driven Approach for the
  Development of Winograd Schemas}. In: Proceedings of the 12th International
  Conference on Agents and Artificial Intelligence - Volume 2: ICAART,. pp.
  26--35. INSTICC, SciTePress (2020). \doi{10.5220/0008902600260035}

\bibitem{joulin2016fasttext}
Joulin, A., Grave, E., Bojanowski, P., Douze, M., J{\'e}gou, H., Mikolov, T.:
  {FastText.zip: Compressing Text Classification Models}. arXiv preprint
  arXiv:1612.03651  (2016)

\bibitem{kocijan2019wikicrem}
Kocijan, V., Camburu, O.M., Cretu, A.M., Yordanov, Y., Blunsom, P.,
  Lukasiewicz, T.: {WikiCREM: A Large Unsupervised Corpus for Coreference
  Resolution}. arXiv preprint arXiv:1908.08025  (2019)

\bibitem{kocijan2019surprisingly}
Kocijan, V., Cretu, A.M., Camburu, O.M., Yordanov, Y., Lukasiewicz, T.: {A
  Surprisingly Robust Trick for Winograd Schema Challenge}. arXiv preprint
  arXiv:1905.06290  (2019)

\bibitem{kocijan2020review}
Kocijan, V., Lukasiewicz, T., Davis, E., Marcus, G., Morgenstern, L.: {A Review
  of Winograd Schema Challenge Datasets and Approaches} (2020)

\bibitem{inproceedingsLe}
Le, T.T.H., Kim, J., Kim, H.: {An Effective Intrusion Detection Classifier
  Using Long Short-Term Memory with Gradient Descent Optimization}. pp.~1--6
  (02 2017). \doi{10.1109/PlatCon.2017.7883684}

\bibitem{lecun2015deep}
LeCun, Y., Bengio, Y., Hinton, G.: {Deep Learning}. {Nature}
  \textbf{521}(7553),  436--444 (2015)

\bibitem{levesque2012winograd}
Levesque, H., Davis, E., Morgenstern, L.: {The Winograd Schema Challenge}. In:
  Thirteenth International Conference on the Principles of Knowledge
  Representation and Reasoning (2012)

\bibitem{levesque2014our}
Levesque, H.J.: {On Our Best Behaviour}. Artificial Intelligence  \textbf{212},
   27--35 (2014)

\bibitem{lin2020tttttackling}
Lin, S.C., Yang, J.H., Nogueira, R., Tsai, M.F., Wang, C.J., Lin, J.:
  {TTTTTackling WinoGrande Schemas} (2020)

\bibitem{liu2004conceptnet}
Liu, H., Singh, P.: {ConceptNet — A Practical Commonsense Reasoning
  Tool-Kit}. BT technology journal  \textbf{22}(4),  211--226 (2004)

\bibitem{liu2016probabilistic}
Liu, Q., Jiang, H., Evdokimov, A., Ling, Z.H., Zhu, X., Wei, S., Hu, Y.:
  {Probabilistic Reasoning via Deep Learning: Neural Association Models}. arXiv
  preprint arXiv:1603.07704  (2016)

\bibitem{liu2019roberta}
Liu, Y., Ott, M., Goyal, N., Du, J., Joshi, M., Chen, D., Levy, O., Lewis, M.,
  Zettlemoyer, L., Stoyanov, V.: {RoBERTa: A Robustly Optimized BERT
  Pretraining Approach} (2019)

\bibitem{michael2009reading}
Michael, L.: {Reading Between the Lines.} In: IJCAI. pp. 1525--1530 (2009)

\bibitem{morgenstern2016planning}
Morgenstern, L., Davis, E., Ortiz, C.L.: {Planning, Executing, and Evaluating
  the Winograd Schema Challenge}. AI Magazine  \textbf{37}(1),  50--54 (2016)

\bibitem{peer2015beyond}
Peer, E., Samat, S., Brandimarte, L., Acquisti, A.: In: Diehl, K.,
  Carolyn~Yoon, D. (eds.) {Beyond the Turk: An Empirical Comparison of
  Alternative Platforms for Crowdsourcing Online Research}. NA - Advances in
  Consumer Research, vol.~43, pp. 18--22. MN : Association for Consumer
  Research (2015)

\bibitem{kn:peng51solving}
Peng, H., Khashabi, D., Roth, D.: {Solving Hard Coreference Problems}. In:
  Proceedings of the 2015 Conference of the North American Chapter of the
  Association for Computational Linguistics: Human Language Technologies. pp.
  809--819 (2015)

\bibitem{pennington2014glove}
Pennington, J., Socher, R., Manning, C.D.: {Glove: Global Vectors for Word
  Representation}. In: Proceedings of the 2014 conference on empirical methods
  in natural language processing (EMNLP). pp. 1532--1543 (2014)

\bibitem{probst2019hyperparameters}
Probst, P., Wright, M.N., Boulesteix, A.L.: {Hyperparameters and Tuning
  Strategies for Random Forest}. Wiley Interdisciplinary Reviews: Data Mining
  and Knowledge Discovery  \textbf{9}(3),  e1301 (2019)

\bibitem{qi2018and}
Qi, Y., Sachan, D.S., Felix, M., Padmanabhan, S.J., Neubig, G.: {When and Why
  Are Pre-Trained Word Embeddings Useful for Neural Machine Translation?} arXiv
  preprint arXiv:1804.06323  (2018)

\bibitem{kn:Rahman:2012:RCC:2390948.2391032}
Rahman, A., Ng, V.: {Resolving Complex Cases of Definite Pronouns: The Winograd
  Schema Challenge}. In: Proceedings of the 2012 Joint Conference on Empirical
  Methods in Natural Language Processing and Computational Natural Language
  Learning. pp. 777--789. EMNLP-CoNLL '12, Association for Computational
  Linguistics, Stroudsburg, PA, USA (2012),
  \url{http://dl.acm.org/citation.cfm?id=2390948.2391032}

\bibitem{ruder2019transfer}
Ruder, S., Peters, M.E., Swayamdipta, S., Wolf, T.: {Transfer Learning in
  Natural Language Processing}. In: Proceedings of the 2019 Conference of the
  North American Chapter of the Association for Computational Linguistics:
  Tutorials. pp. 15--18 (2019)

\bibitem{rudinger-etal-2018-gender}
Rudinger, R., Naradowsky, J., Leonard, B., Van~Durme, B.: {Gender Bias in
  Coreference Resolution}. In: Proceedings of the 2018 Conference of the North
  {A}merican Chapter of the Association for Computational Linguistics: Human
  Language Technologies, Volume 2 (Short Papers). pp. 8--14. Association for
  Computational Linguistics, New Orleans, Louisiana (Jun 2018).
  \doi{10.18653/v1/N18-2002}, \url{https://www.aclweb.org/anthology/N18-2002}

\bibitem{sakaguchi2020winogrande}
Sakaguchi, K., Le~Bras, R., Bhagavatula, C., Choi, Y.: {WinoGrande: An
  Adversarial Winograd Schema Challenge at Scale}. In: Proceedings of the AAAI
  Conference on Artificial Intelligence. vol.~34, pp. 8732--8740 (2020)

\bibitem{sap2019atomic}
Sap, M., Le~Bras, R., Allaway, E., Bhagavatula, C., Lourie, N., Rashkin, H.,
  Roof, B., Smith, N.A., Choi, Y.: {ATOMIC: An Atlas of Machine Commonsense for
  If-Then Reasoning}. In: Proceedings of the AAAI Conference on Artificial
  Intelligence. vol.~33, pp. 3027--3035 (2019)

\bibitem{schmidhuber2015deep}
Schmidhuber, J.: {Deep Learning in Neural Networks: An Overview}. Neural
  networks  \textbf{61},  85--117 (2015)

\bibitem{kn:sharma2015towards}
Sharma, A., Vo, N.H., Aditya, S., Baral, C.: {Towards Addressing the Winograd
  Schema Challenge - Building and Using a Semantic Parser and a Knowledge
  Hunting Module}. In: Proceedings of the Twenty-Fourth International Joint
  Conference on Artificial Intelligence, IJCAI. pp. 25--31 (2015)

\bibitem{singh2002open}
Singh, P., Lin, T., Mueller, E.T., Lim, G., Perkins, T., Zhu, W.L.: {Open Mind
  Common Sense: Knowledge Acquisition From the General Public}. In: OTM
  Confederated International Conferences" On the Move to Meaningful Internet
  Systems". pp. 1223--1237. Springer (2002)

\bibitem{socher2012deep}
Socher, R., Bengio, Y., Manning, C.D.: {Deep learning for NLP (without magic)}.
  In: Tutorial Abstracts of ACL 2012. pp.~5--5. Association for Computational
  Linguistics (2012)

\bibitem{specia2009estimating}
Specia, L., Turchi, M., Cancedda, N., Dymetman, M., Cristianini, N.:
  {Estimating the Sentence-Level Quality of Machine Translation Systems}. In:
  13th Conference of the European Association for Machine Translation. pp.
  28--37 (2009)

\bibitem{speer2017conceptnet}
Speer, R., Chin, J., Havasi, C.: Conceptnet 5.5: An open multilingual graph of
  general knowledge. In: Thirty-First AAAI Conference on Artificial
  Intelligence (2017)

\bibitem{sundermeyer2015feedforward}
Sundermeyer, M., Ney, H., Schl{\"u}ter, R.: {From Feedforward to Recurrent Lstm
  Neural Networks for Language Modeling}. IEEE/ACM Transactions on Audio,
  Speech, and Language Processing  \textbf{23}(3),  517--529 (2015)

\bibitem{sundermeyer2012lstm}
Sundermeyer, M., Schl{\"u}ter, R., Ney, H.: {Lstm Neural Networks for Language
  Modeling}. In: Thirteenth annual conference of the international speech
  communication association (2012)

\bibitem{10.1007/978-3-030-35288-2_18}
Suresh, M., Taib, R., Zhao, Y., Jin, W.: {Sharpening the BLADE: Missing Data
  Imputation Using Supervised Machine Learning}. In: Liu, J., Bailey, J. (eds.)
  AI 2019: Advances in Artificial Intelligence. pp. 215--227. Springer
  International Publishing, Cham (2019)

\bibitem{talmor2018commonsenseqa}
Talmor, A., Herzig, J., Lourie, N., Berant, J.: {CommonsenseqA: A Question
  Answering Challenge Targeting Commonsense Knowledge}. arXiv preprint
  arXiv:1811.00937  (2018)

\bibitem{trinh2018simple}
Trinh, T.H., Le, Q.V.: {A Simple Method for Commonsense Reasoning}. arXiv
  preprint arXiv:1806.02847  (2018)

\bibitem{kn:Valiant:2006knowledge}
Valiant, L.G.: {Knowledge Infusion}. In: Proceedings of the 21st National
  Conference on Artificial Intelligence - Volume 2. pp. 1546--1551. AAAI'06,
  AAAI Press (2006), \url{http://dl.acm.org/citation.cfm?id=1597348.1597438}

\bibitem{wang2019glue}
Wang, A., Singh, A., Michael, J., Hill, F., Levy, O., Bowman, S.R.: {GLUE: A
  Multi-Task Benchmark and Analysis Platform for Natural Language
  Understanding}. In: 7th International Conference on Learning Representations,
  ICLR 2019 (2019)

\bibitem{webster2018mind}
Webster, K., Recasens, M., Axelrod, V., Baldridge, J.: {Mind the Gap: A
  Balanced Corpus of Gendered Ambiguous Pronouns}. Transactions of the
  Association for Computational Linguistics  \textbf{6},  605--617 (2018)

\bibitem{wilson2005opinionfinder}
Wilson, T., Hoffmann, P., Somasundaran, S., Kessler, J., Wiebe, J., Choi, Y.,
  Cardie, C., Riloff, E., Patwardhan, S.: {Opinionfinder: A System for
  Subjectivity Analysis}. In: Proceedings of HLT/EMNLP 2005 Interactive
  Demonstrations. pp. 34--35 (2005)

\bibitem{wilson2005recognizing}
Wilson, T., Wiebe, J., Hoffmann, P.: {Recognizing Contextual Polarity in
  Phrase-Level Sentiment Analysis}. In: Proceedings of Human Language
  Technology Conference and Conference on Empirical Methods in Natural Language
  Processing (2005)

\bibitem{zhao-etal-2018-gender}
Zhao, J., Wang, T., Yatskar, M., Ordonez, V., Chang, K.W.: {Gender Bias in
  Coreference Resolution: Evaluation and Debiasing Methods}. In: Proceedings of
  the 2018 Conference of the North {A}merican Chapter of the Association for
  Computational Linguistics: Human Language Technologies, Volume 2 (Short
  Papers). pp. 15--20. Association for Computational Linguistics, New Orleans,
  Louisiana (Jun 2018). \doi{10.18653/v1/N18-2003},
  \url{https://www.aclweb.org/anthology/N18-2003}

\end{thebibliography}
\end{document}